\documentclass[letterpaper, 10 pt, journal, twoside]{IEEEtran}

\usepackage{wrapfig}
\usepackage{subfigure}
\usepackage{amssymb}
\usepackage{xcolor}
\usepackage{todonotes}
\let\origtodo\todo
\newcommand{\basetodo}{\origtodo}

\renewcommand{\todo}[1]{\basetodo[inline]{#1}}

\usepackage[colorlinks=true,hyperindex=true,bookmarksnumbered,bookmarks=true]{hyperref} 
\usepackage{multirow}
\usepackage{bbm}

\usepackage[colorlinks=true,hyperindex=true,bookmarksnumbered,bookmarks=true]{hyperref} 
\usepackage{multirow}

\usepackage[utf8]{inputenc}
\usepackage[english]{babel}

\usepackage{listings}
\usepackage{color}
\usepackage{tikz}
\usepackage{tikz-layers}
\usepackage{array}
\newcolumntype{C}[1]{>{\centering\arraybackslash}p{#1}}
\usetikzlibrary{shapes.geometric,automata,positioning,arrows,shadows,,backgrounds,calc}

\definecolor{frameColor}{RGB}{128,128,128}
\definecolor{initialStateColor}{RGB}{10,100,10}
\definecolor{impactStateColor}{RGB}{0,255,0}
\definecolor{admittanceStateColor}{RGB}{0,0,255}

\definecolor{resetStateColor}{RGB}{10,100,10}

\definecolor{stateColor}{RGB}{51,204,204}

\definecolor{controllerColor}{RGB}{10,10,255}
\definecolor{plannerColor}{RGB}{10,150,150}
\definecolor{estimatorColor}{RGB}{10,200,10}
\definecolor{robotColor}{RGB}{255,0,0}
\definecolor{modelColor}{RGB}{180,100,10}

\usepackage{kantlipsum}

\usepackage{setspace}

\usepackage{listings}
\newcommand{\quickEq}[2]{
  \begin{equation}
    \label{#1}
    {#2}
  \end{equation}
}

\usepackage{xspace}
\newcommand{\unitHz}{~Hz\xspace}
\newcommand{\unitMs}{~ms\xspace}

\newcommand{\unitPosJS}{~rad\xspace}
\newcommand{\unitVelTS}{~m/s\xspace}
\newcommand{\unitVelJS}{~rad/s\xspace}

\usepackage{pifont}
\newcommand{\contactVel}{\bs{v}}
\newcommand{\nContactVel}{v_n}
\newcommand{\xContactVel}{v_x}
\newcommand{\yContactVel}{v_y}

\newcommand{\iim}{W}
\newcommand{\iimcrb}{\iim_{\text{crb}}}
\newcommand{\iimgm}{\iim_{\text{gm}}}

\newcommand{\postImpact}[1]{{#1}^{+}}
\newcommand{\preImpact}[1]{#1^{-}}
\newcommand{\coefR}{c_{\text{r}}}
\newcommand{\coefF}{\mu}

\newcommand{\tp}[1]{{#1}_{\perp}}

\newcommand{\nImpulse}{{\impulseScalar_{n}}}

\definecolor{csrStateColor}{RGB}{255,255,153}
\definecolor{scrStateColor}{RGB}{153,204,0}
\definecolor{crStateColor}{RGB}{153,204,155}

\usepackage{stackengine}

\newcommand{\body}[1]{#1^{b}}
\newcommand{\spatial}[1]{#1^{s}}

\newcommand{\inertialFrame}{O}

\newcommand{\abs}[1]{\left\vert#1\right\vert}

\newcommand{\dualCross}[2]{ #1 \times^* #2 }

\newcommand{\wrenchAS}[2]{S}

\usepackage{algorithmicx}
\usepackage{algorithm}
\usepackage{algpseudocode}

\newcommand{\crbGInertia}{
  {^{\text{crb}}I}
}

\newcommand{\dof}{D}
\newcommand{\inertiaMatrix}{M}
\newcommand{\gravityandcoriolis}{\mathbf{N}}

\newcommand{\eInertiaMatrix}{{I_{\text{eq}}}}

\newcommand{\agg}[3]{
\sum^{#2}_{#1=1}{#3}
}

\newcommand{\vc}[1]{\bs{#1}}

\newcommand{\vectorTwo}[2]{
  \begin{bmatrix}
    #1\\
    #2
\end{bmatrix}
}

\newcommand{\vectorThree}[3]{
  \begin{bmatrix}
    #1\\
    #2\\
    #3
\end{bmatrix}
}

\newcommand{\matrixTwo}[4]{
  \begin{bmatrix}
    #1 & #2\\
    #3 & #4
\end{bmatrix}
}

\newcommand{\matrixThree}[9]{
  \begin{bmatrix}
    #1 & #2 & #3\\
    #4 & #5 & #6\\
    #7 & #8 & #9
\end{bmatrix}
}

\newcommand{\bodyVel}[2]{
  \textcolor{bodyVelColor}{
    \vc{V}_{#1#2}
  }}

\newcommand{\bodyTV}[2]{
  \textcolor{bodyVelColor}{
    \vc{v}_{#1#2}
  }}

\newcommand{\rotation}[2]{R_{#1 #2}}

\newcommand{\rotationInv}[2]{R^{\top}_{#1 #2}}

\newcommand{\translation}[2]{
  \vc{p}_{#1 #2}
}

\newcommand{\translationSkew}[2]{\skewMatrix{\vc{p}}_{#1 #2}}

\usepackage{dsfont}
\newcommand{\identityMatrix}{\mathds{1}} 
\newcommand{\zeroMatrix}{0} 

\newcommand{\twistTransformTwo}[2]{
  \textcolor{geometricColorVariable}{
    \adg{#2}{#1}
  }
}

\newcommand{\twistTransformTwoDef}[2]{
  \textcolor{geometricColorVariable}{
    \adgDef{#2}{#1}
  }
}

\newcommand{\twistTransform}[2]{
  \textcolor{geometricColorVariable}{
    \adgInv{#1}{#2}
  }
}

\newcommand{\twistTransformDef}[2]{
  \adgInvDef{#1}{#2}
}

\newcommand{\bodyVelTransform}[3]{
  \textcolor{bodyVelColor}{
    \twistTransform{#1}{#2}\bodyVel{#3}{#1} + \bodyVel{#1}{#2}
  }
}

\newcommand{\adg}[2]{
  \textcolor{geometricColorVariable}{
    Ad_{g_{#1#2}}
  }
}

\newcommand{\adgDef}[2]{
  \textcolor{geometricColorVariable}{
  \begin{bmatrix}
    \rotation{#1}{#2} & \translationSkew{#1}{#2}\rotation{#1}{#2} \\
    \zeroMatrix & \rotation{#1}{#2}
  \end{bmatrix}
}}

\newcommand{\adgInv}[2]{
  \textcolor{geometricColorVariable}{
  Ad^{-1}_{g_{#1#2}}
}}

\newcommand{\adgTrans}[2]{
  \textcolor{geometricColorVariable}{
  Ad^{\top}_{g_{#1#2}}
}}
\newcommand{\adgTransDef}[2]{
  \textcolor{geometricColorVariable}{
  \begin{bmatrix}
    \rotationInv{#1}{#2} & \zeroMatrix \\
     -\rotationInv{#1}{#2}\translationSkew{#1}{#2} & \rotationInv{#1}{#2}
\end{bmatrix}
}}

\newcommand{\adgInvDef}[2]{
  \textcolor{geometricColorVariable}{
  \begin{bmatrix}
    \rotationInv{#1}{#2} & -\rotationInv{#1}{#2}\translationSkew{#1}{#2} \\
    \zeroMatrix & \rotationInv{#1}{#2}
\end{bmatrix}
}}

\newcommand{\geometricFTDef}[2]{
  \textcolor{geometricColorVariable}{
    \adgInvTransDef{#2}{#1}
  }
}

\newcommand{\geometricFTTwo}[2]{
  \textcolor{geometricColorVariable}{
    \adgTrans{#1}{#2}
  }
}

\newcommand{\geometricFT}[2]{
  \textcolor{geometricColorVariable}{
    \adgInvTrans{#2}{#1}
  }
}

\newcommand{\adgInvTrans}[2]{
  \textcolor{geometricColorVariable}{
  Ad^{\top}_{g^{-1}_{#1#2}}
}}

\newcommand{\adgInvTransDef}[2]{
  \textcolor{geometricColorVariable}{
  \begin{bmatrix}
    \rotation{#1}{#2} & \zeroMatrix \\
    \translationSkew{#1}{#2}\rotation{#1}{#2}  & \rotation{#1}{#2}
  \end{bmatrix}
  }
}

\newcommand{\wrench}{\bs{W}}

\newcommand{\bodyGInertia}[2]{
\textcolor{bodyVelColor}{
  \body{\gInertia}_{#1 #2}
  }
}

\newcommand{\spatialGInertia}[2]{
  \textcolor{spatialVelColor}{
    \spatial{\gInertia}_{#1#2}
  }
}

\newcommand{\metaInertia}{\mathcal{I}}

\newcommand{\spatialMetaInertia}[2]
{
\textcolor{spatialVelColor}{ 
  \spatial{\metaInertia}_{{#1}{#2}}
}
}

\newcommand{\gInertia}{\mathcal{M}}

\newcommand{\measured}[1]{#1^{\circ}}

\newcommand{\aError}[1]{\bar{\bs{e}}_{#1}}
\newcommand{\error}[1]{\bs{e}_{#1}}

\newcommand{\complexity}[1]{
  {\mathcal{O}(#1)}
}

\newcommand{\bs}{\boldsymbol}

\newcommand{\skewMatrix}[1]{\widehat{#1}}

\newcommand{\skewMatrixTwo}[1]{({#1})^{\widehat{~}}}

\newcommand{\transpose}[1]{{#1}^\top}

\newcommand{\pseudoInverseRowDef}[1]{\transpose{#1}\inverse{( #1 \transpose{#1} )}}

\newcommand{\inverse}[1]{{#1}^{-1}}

\newcommand{\cframe}[1]{\mathcal{F}_{#1}}

\newcommand{\mass}{\text{m}}

\newcommand{\contactPoint}{\bs{p}}

\newcommand{\force}{\bs{f}}

\newcommand{\torque}{\bs{\tau}}

\newcommand{\impactDuration}{\delta t}

\newcommand{\jump}{\Delta}

\newcommand{\com}{{\bs{c}}}

\newcommand{\impulse}{\bs{\iota}} 
\newcommand{\impulseEstimate}{\bs{\iota}^{\circ}} 
\newcommand{\impulseScalar}{\iota} 

\newcommand{\jangles}{\bs{q}}
\newcommand{\jvelocities}{\dot{\bs{q}}}
\newcommand{\jaccelerations}{\ddot{\bs{q}}}

\newcommand{\jacobian}{J}

\newcommand{\jvelocitiesJumpEstimate}{\jump \jvelocities^{\circ}}

\newtheorem{theorem}{Theorem}

\newtheorem{remark}{Remark}[section]

\newtheorem{Example}[theorem]{Example}

\usepackage{mathtools}

\newcommand{\RRv}[1]{\mathbb{R}^{#1}}
\newcommand{\RRm}[2]{\mathbb{R}^{#1 \times #2}}

\definecolor{dkgreen}{rgb}{0,0.6,0}
\definecolor{gray}{rgb}{0.5,0.5,0.5}
\definecolor{mauve}{rgb}{0.58,0,0.82}

\NewDocumentCommand{\cpp}{v}{%
\texttt{\textcolor{blue}{#1}}%
}

\NewDocumentCommand{\secRef}{v}{%
Sec.~\ref{#1}}

\NewDocumentCommand{\remarkRef}{v}{%
Remark.~\ref{#1}}

\NewDocumentCommand{\tableRef}{v}{%
Table.~\ref{#1}}

\NewDocumentCommand{\algRef}{v}{%
Algorithm.~\ref{#1}}

\NewDocumentCommand{\lemmaRef}{v}{%
Lemma.~\ref{#1}}

\NewDocumentCommand{\theoremRef}{v}{%
Theorem.~\ref{#1}}

\NewDocumentCommand{\figRef}{v}{%
  Fig.~\ref{#1}}

\NewDocumentCommand{\appRef}{v}{%
  Appendix~\ref{#1}}

\NewDocumentCommand{\lineRef}{v}{%
  line.~\ref{#1}}

\NewDocumentCommand{\probRef}{v}{%
Problem~\ref{#1}}

\NewDocumentCommand\orderedTwoS{mm}%
  {$<$#1,#2$>$}
\NewDocumentCommand\orderedThreeS{mmm}%
  {$<$#1,#2,#3$>$}

\definecolor{svaColor}{RGB}{7, 160, 2}
\definecolor{uncertainColor}{RGB}{255, 0, 0}

\definecolor{spatialVelColor}{RGB}{189, 38, 96}
\definecolor{geometricColor}{RGB}{66, 126, 245}
\definecolor{bodyVelColor}{RGB}{13, 191, 191}

\colorlet{svaColorVariable}{svaColor}
\colorlet{geometricColorVariable}{geometricColor}

\newif\ifBlockComment
\BlockCommentfalse 

\newcommand{\solution}{\jump \jvelocities^*}
\newcommand{\proposedSolution}{\jump \jvelocities^*_{\text{crb}}}

\newcommand{\classicalSolution}{\jump \jvelocities^*_{\text{c}}}
\newcommand{\gmSolution}{\jump \jvelocities^*_{\text{gm}}}

\newcommand{\crbImpulse}{ \impulse^*_{\text{crb}}}
\newcommand{\gmImpulse}{\impulse^*_{\text{gm}}}
\newcommand{\classicalImpulse}{\impulse_{\text{c}}}

\newcommand{\measuredImpulse}{ \impulseEstimate}

\newcommand{\moment}{\bs{m}}
\newcommand{\rVel}{\bs{w}}

\usepackage{caption}
\DeclareMathOperator*{\argmin}{arg\,min} 

\begin{document}
\newenvironment{renum}
{\begin{enumerate}\renewcommand\labelenumi{ C.\arabic{enumi}}}
  {\end{enumerate}}

\def\quote#1{``#1''}


\title{Predicting Impact-Induced Joint Velocity Jumps on Kinematic-Controlled Manipulator}

\author{Yuquan Wang$^{1}$, Niels Dehio$^{1}$, and Abderrahmane Kheddar$^{1,2}$,~\IEEEmembership{Fellow,~IEEE}%
\thanks{
  This work is in part supported by the Research Project I.AM. through the European Union H2020 program (GA 871899).} 
\thanks{$^{1}$ Y. Wang, N. Dehio and A. Kheddar are with the CNRS-University of Montpellier LIRMM, Montpellier, 34090, France. 
}%
\thanks{Corresponding author: Y. WANG. yuquan.wang@lirmm.fr}%
\thanks{$^{2}$ A. Kheddar is also with the CNRS-AIST Joint Robotics Laboratory, IRL, Tsukuba, Japan.}%
}

\maketitle


\begin{abstract}
  In order to enable on-purpose robotic impact tasks, predicting joint-velocity jumps is essential to enforce controller feasibility and hardware integrity.
  We observe a considerable prediction error of a commonly-used approach in robotics compared against 250 benchmark experiments with the Panda manipulator. 
  We reduce the average prediction error by 81.98$\%$ as follows: First, we focus on task-space equations without inverting the ill-conditioned joint-space inertia matrix.
  Second, before the impact event, we compute the equivalent inertial properties of the end-effector tip
  considering that a high-gains (stiff) kinematic-controlled manipulator behaves like a composite-rigid body. 
\end{abstract}


\IEEEpeerreviewmaketitle

\section{Introduction}
\IEEEPARstart{W}{e} aim at enabling robots to impact with the highest \emph{feasible} velocity\footnote{Refer to Impact-Aware Manipulation \url{https://i-am-project.eu}}.
An impact-aware robot controller must embed predicted post-impact joint velocities to prevent violation of hardware limitations~\cite{wang2019rss}.
To our best knowledge, existing  prediction approaches are not well-assessed with a rich ground-truth, e.g., based on large impact-data collected from high-gain joint-velocity (or -position) controlled robots, we refer to as \emph{kinematic controlled robots}.

\begin{figure}[tbp!]
  \vspace{-3mm}
  \centering
  \includegraphics[width=\columnwidth]{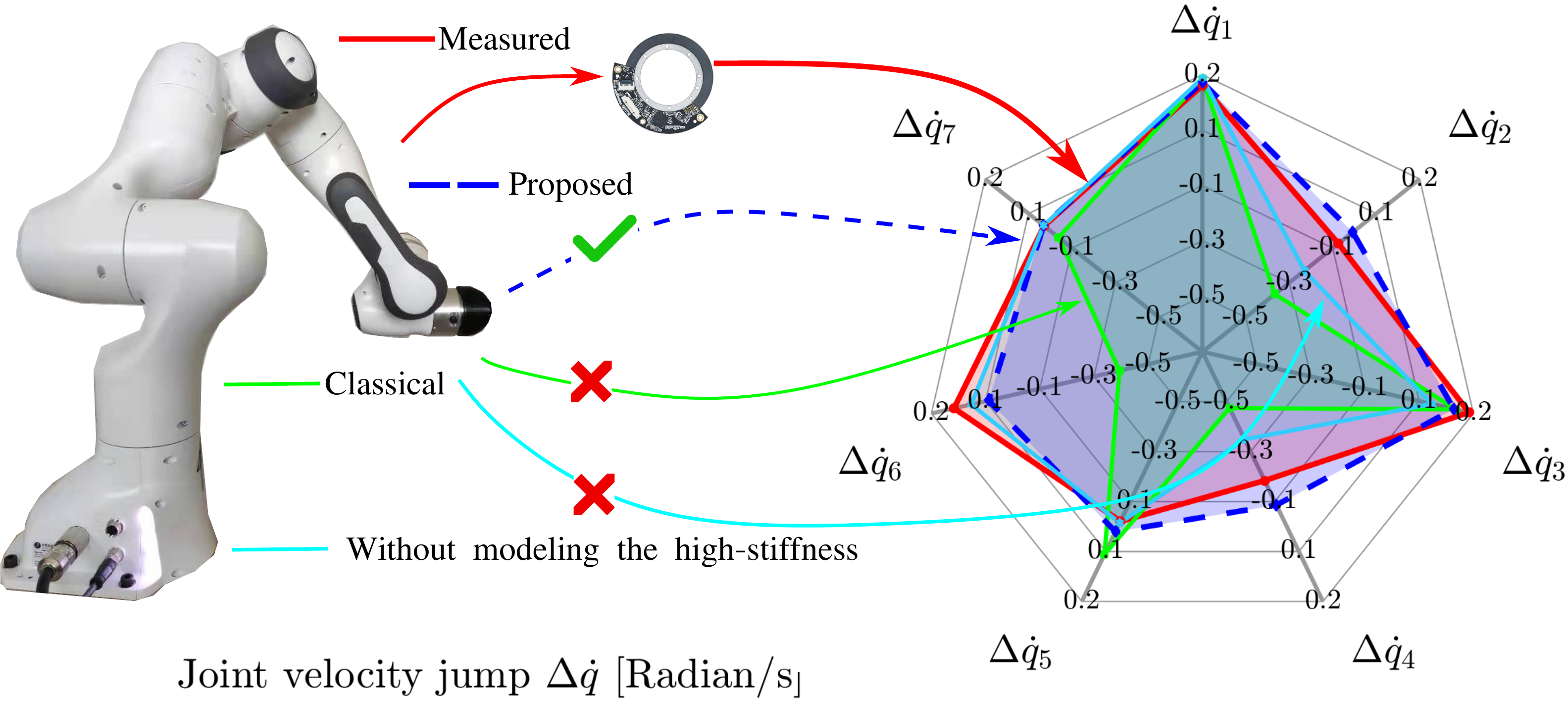}
  \caption{
    The predictions following~\cite{zheng1985mathematical} (green polygon) are substantially different from the measured joint velocity jumps $\jump \jvelocities$ (red), which is defined in Sec.~\ref{sec:measure_qd_jump}. 
    Computation without modeling high-stiffness aspects~\cite{lankarani2000poisson} also leads to significant errors (cyan).
    Our proposed prediction has the smallest error (dashed-blue). 
}
\label{fig:comparison}
\vspace{-6mm}
\end{figure}

The methodology proposed in the late 1980s~\cite{zheng1985mathematical} predicts the impact-induced joint-velocity jumps as the product of the inverted \emph{joint-space inertia matrix} (JSIM) and the joint-space impulse, see also~\cite{grizzle2014automatica,siciliano2016springer}. Unfortunately, this prediction scheme poorly matches the measurements collected from our benchmark experiments, see~\figRef{fig:comparison}. 
We summarize three reasons that potentially explain such discrepancies:\\
\textbf{R1)}~\emph{Inaccurate impulse calculation}:
accurate task-space impulse calculation\footnote{We reserve \emph{prediction} for the concept: \emph{predicting the joint velocity jumps}. Thus, we use \emph{impulse calculation} to describe \emph{impulse prediction}.} is essential for solving the joint velocity jumps from momentum conservation; i.e., the momentum change  is equal to the impulse. The classical approach relies on inaccurate calculation~\cite[Fig.~9]{wang2022ral}, which  overestimates the normal impulse and does not account for contact friction effects~\cite[Eq.~4.8]{stronge2000book},~\cite{jia2017ijrr}.\\
\textbf{R2)}~\emph{Ill-conditioned JSIM}: 
The computation is sensitive to the impulse-calculation error since it inverts the ill-conditioned JSIM.
Featherstone~\cite{featherstone2008book} summarized that the condition number of the JSIM grows within the range of $\complexity{N}$ and $\complexity{N^4}$ for a serial robotic manipulator with $N$ being the number of revolute joints. Therefore, the inverted JSIM amplifies errors associated with the impulse \cite[Chapter~10.1]{featherstone2008book}.\\
\textbf{R3)}~\emph{Unmodeled joint-stiffness behavior}:
Standard robot manipulators are mechanically stiff (not backdrivable) and kinematic-controlled to precisely track reference trajectories~\cite[Chapter~9]{siciliano2016springer}.

The joint velocity jump prediction in~\cite{zheng1985mathematical,grizzle2014automatica,siciliano2016springer} does not account for either mechanical non-backdrivability or high-gain PD control (quasi-locked joints) in drivable ones. Both induces rather light-flexibilities of the joints as acknowledged in~\cite{youcef1994impact,ferretti1998impact}, and recently in~\cite[Fig.~9]{wang2022ral}. 

Since three-dimensional impact dynamics models are not well-assessed for articulated robots~\cite{stronge2000book,jia2017ijrr}, we use the measured impulses, or compute the normal impulse with our improved calculation~\cite{wang2022ral} when the tangential impulse is negligible (\textbf{R1}).
We formulate the momentum conservation in the \emph{task-space}.
Hence, our approach does not involve the joint-space formulation relying on the JSIM (\textbf{R2}). 
We modify the task-space impulse-to-velocity mapping considering the joints to be stiff at impact and therefore behaving as a composite-rigid body (CRB) (\textbf{R3}), yet at a time flexible to allow potential post-impact contact mode as sliding/sticking. 

We conducted 250 impact experiments with the seven degrees of freedom Panda manipulator for evaluation. 
The dataset comprises different impact configurations, various normal and tangential contact velocities. We repeat
each condition ten times to eliminate the outliers. The proposed prediction:
\begin{itemize}
\item outperforms the classical approach, reducing the average prediction error by $81.98\%$;
\item is analytical, i.e., does not require the measured impulse. The average prediction error remains on the same level whether the impulse is calculated or measured.
\end{itemize}
Interested readers can reproduce our results with the post-processed data and the MATLAB scripts\footnote{\url{https://gite.lirmm.fr/yuquan/fidynamics/-/wikis/Predicting-Impact-Induced-Sudden-Change-of-Joint-Velocities}}.

\section{Related work}
\label{sec:related_work}
Impacts affect the robot joint velocities almost instantly~\cite{zheng1985mathematical}.
Preemptive actions are necessary to avoid violating the hardware-dependent joint velocity bounds. 
Unlike specifying pre- and post-impact
velocity references in joint space~\cite{rijnen2017icra} or planning impacts carefully~\cite{Stouraitis2020IROS},
it is generally more convenient to formulate quadratic optimization problems in task-space to meet multiple control objectives and constraints \cite{bouyarmane2019tro}.
We preliminarily enabled on-purpose impacts to second order dynamic QP control schemes~\cite{wang2019rss,yuquan2020ijrr}.

Most of the  impact dynamics models focus on the velocity jumps in Cartesian space~\cite{stronge2000book,jia2019ijrr,faik2000modeling}, e.g., batting a flying object~\cite{jia2019ijrr}. 
Solving the joint velocity jumps merely serves as an intermediate step concerning the momentum conservation in joint-space \cite{zheng1985mathematical,lankarani2000poisson,khulief2013modeling}.
Our literature review revealed that the algebraic equations for the \emph{joint-space velocity jumps} proposed by~\cite{zheng1985mathematical} more than 30 years ago are still considered state-of-the-art, see e.g., the robotics textbook~\cite{siciliano2016springer} and the survey by~\cite{grizzle2014automatica}. They are used recently in~\cite{aouaj2021icra}, to extract the rigid impact component in the post-impact joint-velocity oscillatory-behavior (a corotational impact approach).

However, to the authors' best knowledge, the prediction of joint-space velocity jumps has not been confronted to ground-truth experiments
with manipulators that are kinematic-controlled. Also, our former QP controller extension was assessed in simulation only~\cite{wang2019rss}.

Hereafter, we list some issues of the usual computation:

\paragraph{Impulse calculation}
Integrating an impact dynamics model into QP controllers is not trivial. Formulating optimization problems requires closed-form constraints and objectives that are not possible with friction.
There exist analytical solutions only for the planar impacts \cite{khulief2013modeling}.
When the variables are constrained on a plane, the analytical solution exists because the post-impact contact mode is limited to rebounce, sliding, reverse sliding, and sticking \cite{stronge2000book,khulief2013modeling,jia2019ijrr}.
Without a numerical process, it is not possible to determine the tangential velocities for three-dimensional impacts~\cite{stronge2000book,jia2017ijrr}. 

\paragraph{Ill-conditioning issue}
The principle of momentum conservation is often specified in joint space, relying on the joint-space inertia matrix (JSIM).
The JSIM inversion~\cite{zheng1985mathematical} amplifies the impulse calculation
errors ~\cite[Chapter~10.1]{featherstone2008book} due to the well-known ill-conditioning issue of the JSIM  \cite{featherstone2004ijrr}.
Real-robot experiments we conducted proved that incorporating the JSIM inversion for dynamically consistent null-space projections does not significantly show, in practice, theoretical-conceptual superiority~\cite{DietrichIJRR2015,fonseca2021task}. That is why a recent line of research proposed alternative orthogonal null-space projectors that do not rely on the inverted JSIM~\cite{Dehio2021IJRR}.

\paragraph{Unmodeled joint-stiffness behavior}
Early works since the 1990s focus on contact-force modeling for one degree-of-freedom set-up~\cite{youcef1994impact}, or control designs to damp the post-impact oscillations~\cite{ferretti1998impact,xu2000experimental} using compliant models. Textbook impact mechanics usually assume impacts between two free-flying objects~\cite{stronge2000book,jia2017ijrr}.
Unfortunately, most existing joint-velocity-jump predictions do not explicitly account for the joint stiffness at impact, or even consider under-actuated pendulums~\cite{lankarani2000poisson,ganguly2020ijrr}. We preliminarily concluded that the CRB assumption is essential for
task-space velocity-to-impulse calculation for a kinematic-controlled robot~\cite{wang2022ral}. Otherwise, merely fulfilling the joint motion constraint~\cite{lankarani2000poisson,khulief2013modeling} underestimates the equivalent momentum at the manipulator contact point~\cite[Fig.~9]{wang2022ral}. 
We therefore adopt the CRB assumption from~\cite{wang2022ral} for the task-space impulse-to-velocity mapping. 

\section{Background}
\label{sec:problem}

The well-known equations of motion (EOM) for a fixed-base manipulator with $\dof$ degrees of freedom are given as
$\inertiaMatrix \jaccelerations + \gravityandcoriolis = \transpose{\jacobian} \force + \torque$, 
where $\gravityandcoriolis \in \RRv{\dof}$ gathers the centrifugal, Coriolis, and gravitational forces, 
$\torque \in \RRv{\dof}$ denotes the joint actuation torques and $\force \in \RRv{3}$ denotes the external force applied on a contact point.
We denote the positive-definite JSIM as $\inertiaMatrix\in \RRm{\dof}{\dof}$, 
and the translation (linear) part of the  contact point Jacobian as $\jacobian \in \RRm{3}{\dof}$.

  \textbf{Problem statement}: How to accurately predict the impact-induced joint velocity jumps $\jump \jvelocities^*$,
  given prior knowledge of the (pre-)impact robot configuration, i.e., (i) the joint positions $\jangles$, velocities $\jvelocities$, (ii) the equations of motion, (iii) the contact point location with its associated $\jacobian$ and (iv) an impact dynamics model that predicts the task-space impulse $\impulse \in \RRv{3}$
  \quickEq{eq:impulse_contactVel}{
    \impulse = \text{Impact dynamics model}(\jangles, \jump \contactVel),
  }
  where we compute the task-space velocity jump $\jump \contactVel \in \RRv{3}$ 
  based on the coefficient of restitution $\coefR \in [0, 1]$ and the pre-impact velocity $\preImpact{\contactVel}$:
$$
\jump \contactVel = - (1 + \coefR)\preImpact{\contactVel}.
$$
We assume that:
  \begin{enumerate}
  \item \label{assumption1} The impact force is dominant relatively to other forces. Consequently, generalized forces such as the centrifugal forces or motor command torques during impact are negligible~\cite[Chapter~8.1.1]{stronge2000book}. 
  \item \label{assumption2} The  contact area is tiny compared to the robot dimensions  such that a point contact model is appropriate \cite{chatterjee1998new}.
  \item \label{assumption3} The impact induces significant impulsive contact forces and  negligible contact moments \cite{stronge2000book,chatterjee1998new}.
  \item \label{assumption4} The robot manipulator is fully actuated under kinematic control with high gains.
  \item \label{assumption5} The manipulator has a minimum of three degrees of freedom. It impacts at the end-effector, and the joint configuration at impact is not in a singular configuration.
  \end{enumerate}

  In the rest of the paper, we represent the contact velocity $\contactVel$ and the impulse $\impulse$ in
  the contact frame $\cframe{\contactPoint}$. 
  The origin of $\cframe{\contactPoint}$ locates at the contact point. The $z$-axis aligns with the impact normal.  The $x$-axis and $y$-axis span the tangential plane.  

  \begin{remark}
  We handle velocity and wrench transforms in line with the \emph{geometric approach} stated in~\cite{murray1994book}. The notation \emph{Ad} is the abbreviation of \emph{adjoint transform} \cite[Lemma 2.13 on page 56]{murray1994book}, which is associated with rigid body motion $g \in SE(3)$.    We use $Ad_g$ to transform \emph{twists} from one coordinate to another. For instance,  we  transform the wrench $\wrench_{\contactPoint} \in \RRv{6}$ measured from the contact point frame $\cframe{\contactPoint}$ to the centroidal frame $\cframe{\com}$:
  $$
  \wrench_{\com} =  \geometricFTTwo{\contactPoint}{\com}\wrench_{\contactPoint} = \geometricFT{\contactPoint}{\com}\wrench_{\contactPoint} = \geometricFTDef{\contactPoint}{\com}\wrench_{\contactPoint}.
  $$
  \end{remark}

\subsection{The usual approach}

Integrating the EOM
over the impact duration $\impactDuration$ leads to the impulse-momentum conservation in joint space:
\quickEq{eq:law}{
\inertiaMatrix \jump \jvelocities = \transpose{\jacobian}\impulse
}
where $\impulse\in\RRv{3}$ is  $\int^{\impactDuration}_0\force dt$. 
We have $\jump\torque=0$, $\jump \gravityandcoriolis = 0$ due to assumption \ref{assumption1}.
The  usual approach \cite{zheng1985mathematical,grizzle2014automatica,siciliano2016springer} predicts $\jump \jvelocities$ by left multiplying \eqref{eq:law} with  $\inertiaMatrix^{-1}$:
\quickEq{eq:dq_jump_one}{
  \jump \jvelocities = \inverse{\inertiaMatrix} \transpose{\jacobian}\impulse.
}
which is the product of $\inverse{\inertiaMatrix}$ and the joint-space impulse $\transpose{\jacobian}\impulse$. 
The impulse $\impulse$ in \eqref{eq:dq_jump_one} is unknown.
Left multiplying \eqref{eq:dq_jump_one} by  $\jacobian$ yields:
\quickEq{eq:generalized_momentum}{
  \jacobian\jump\jvelocities = \jump \contactVel = {\jacobian \inverse{\inertiaMatrix} \transpose{\jacobian}}\impulse,
}
where the product ${\jacobian \inverse{\inertiaMatrix} \transpose{\jacobian}}$ is positive-definite and invertible due to assumption~\ref{assumption5}.  
The classical approach predicts the task-space impulse as \cite{zheng1985mathematical,grizzle2014automatica,siciliano2016springer}:
\quickEq{eq:fl_impulse}{
  \classicalImpulse =
  \inverse{(\jacobian \inverse{\inertiaMatrix} \transpose{\jacobian})}
  \jump \contactVel.
}
Substituting the calculated impulse \eqref{eq:fl_impulse} into \eqref{eq:dq_jump_one}, the predicted joint velocity jump is:
\quickEq{eq:textbook_dq_jump}{
  \classicalSolution =
  \inverse{\inertiaMatrix} \transpose{\jacobian} \classicalImpulse = 
  \inverse{\inertiaMatrix} \transpose{\jacobian}\inverse{(\jacobian \inverse{\inertiaMatrix} \transpose{\jacobian})}\jump \contactVel.
}
which is equivalently written in a weighted least-squares form:
\quickEq{eq:joint-space-solution}{
  \classicalSolution = \argmin_{\jump \jvelocities} \|\jacobian \jump \jvelocities - \jump \contactVel \|^2_\inertiaMatrix.
}
\begin{figure}[htbp!]
\vspace{-3mm}
\begin{tabular}{C{.19\textwidth}C{.24\textwidth}}
\subfigure [Configuration one] {
    \resizebox{0.17\textwidth}{!}{%
  \begin{tikzpicture}
    \label{fig:contact_posture_one}
    \begin{scope}
      \node[anchor=south west,inner sep=0] (image) at (0,0) {
        \includegraphics[width=0.17\textwidth]{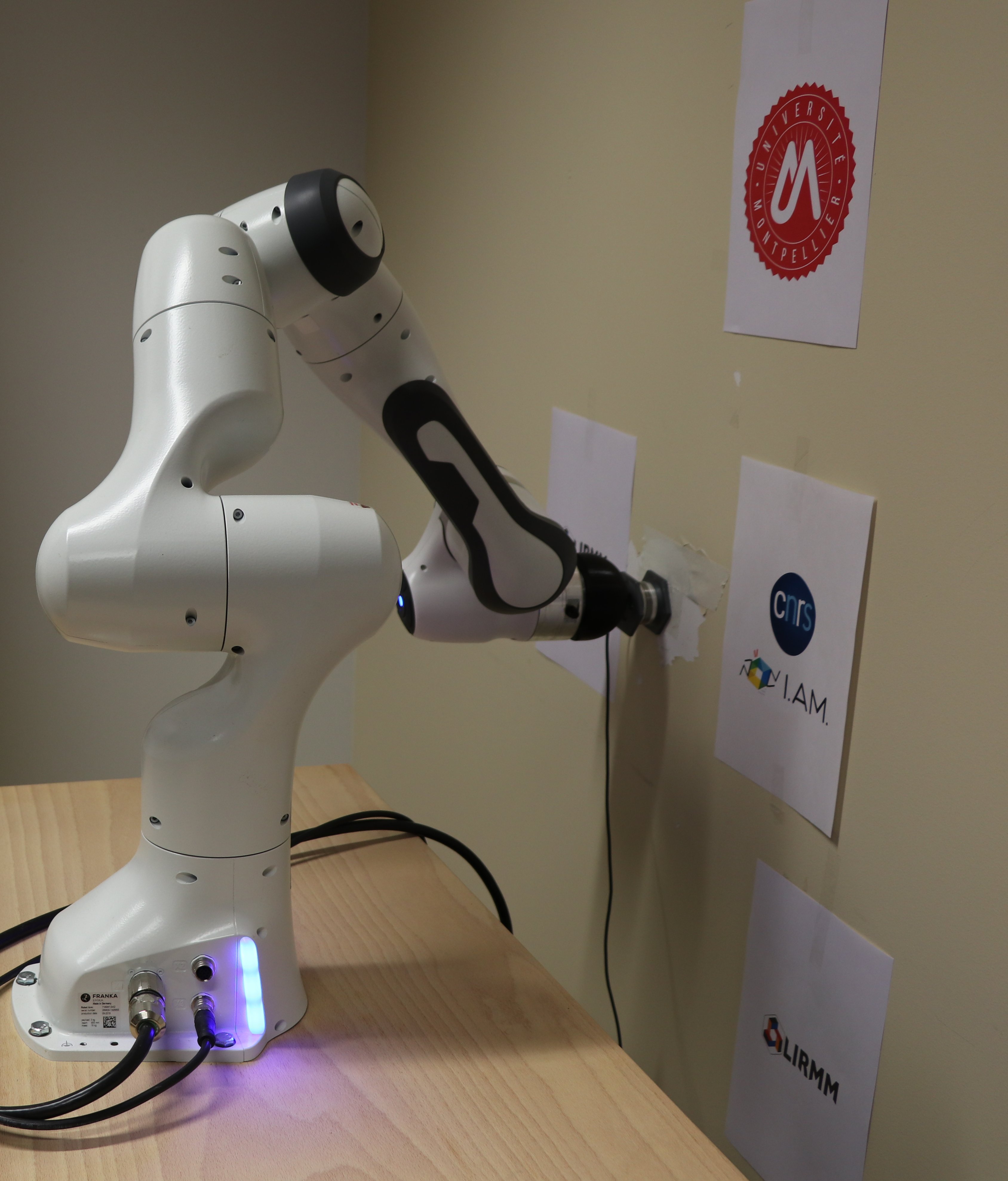}};
    \end{scope}
    \end{tikzpicture} 
    }
} & 
\subfigure [The measured $\jump \jvelocities$] {
    \resizebox{0.25\textwidth}{!}{%
    \begin{tikzpicture}
      \label{fig:measured_qd_jump_one}
    \begin{scope}
      \node[anchor=south west,inner sep=0] (image) at (0,0) {
        \includegraphics[width=0.24\textwidth]{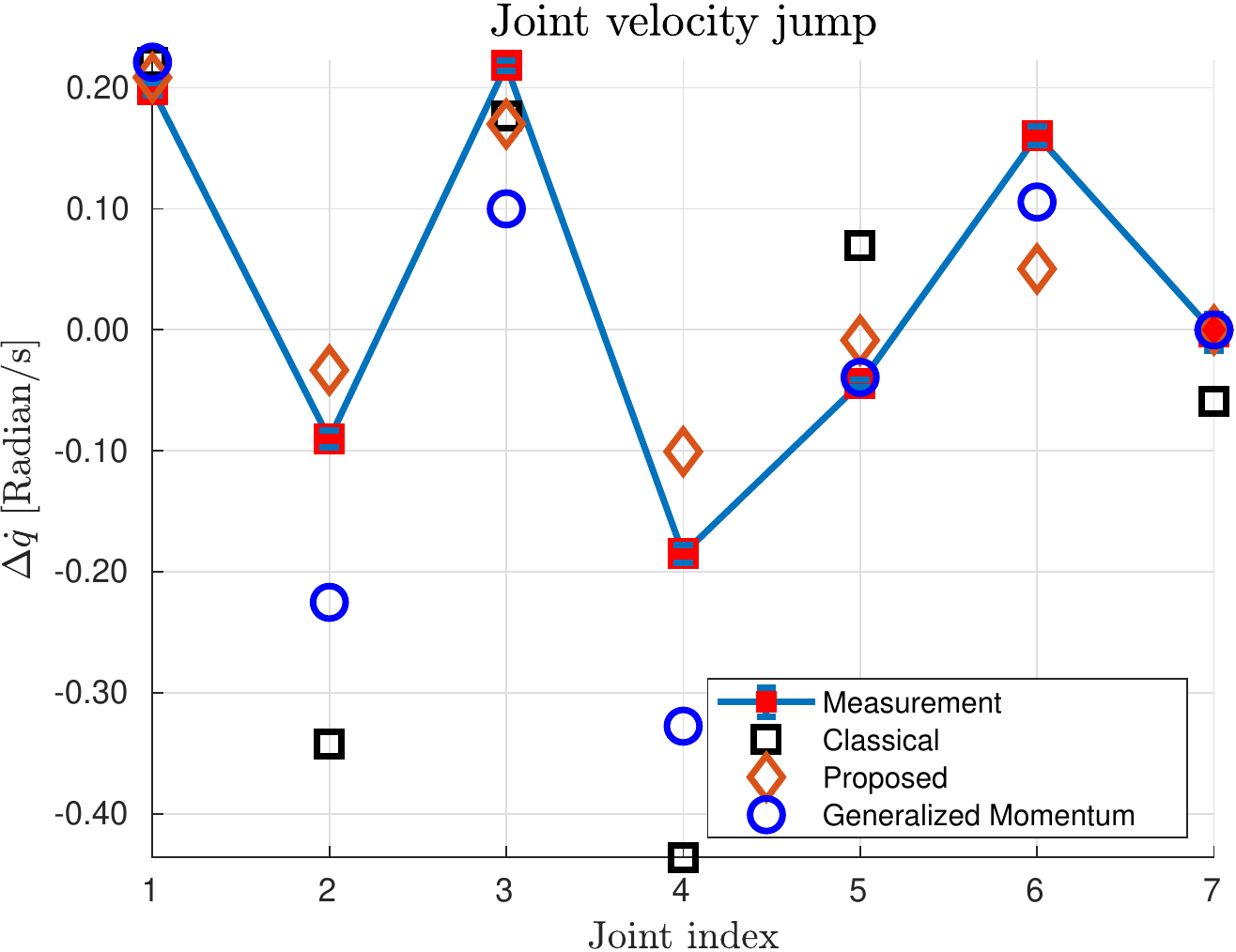}
      };
    \end{scope}
    \end{tikzpicture} 
    }
    } \\
\subfigure [Configuration two] {
    \resizebox{0.17\textwidth}{!}{%
  \begin{tikzpicture}
    \label{fig:contact_posture_two}
    \begin{scope}
      \node[anchor=south west,inner sep=0] (image) at (0,0) {
        \includegraphics[width=0.19\textwidth]{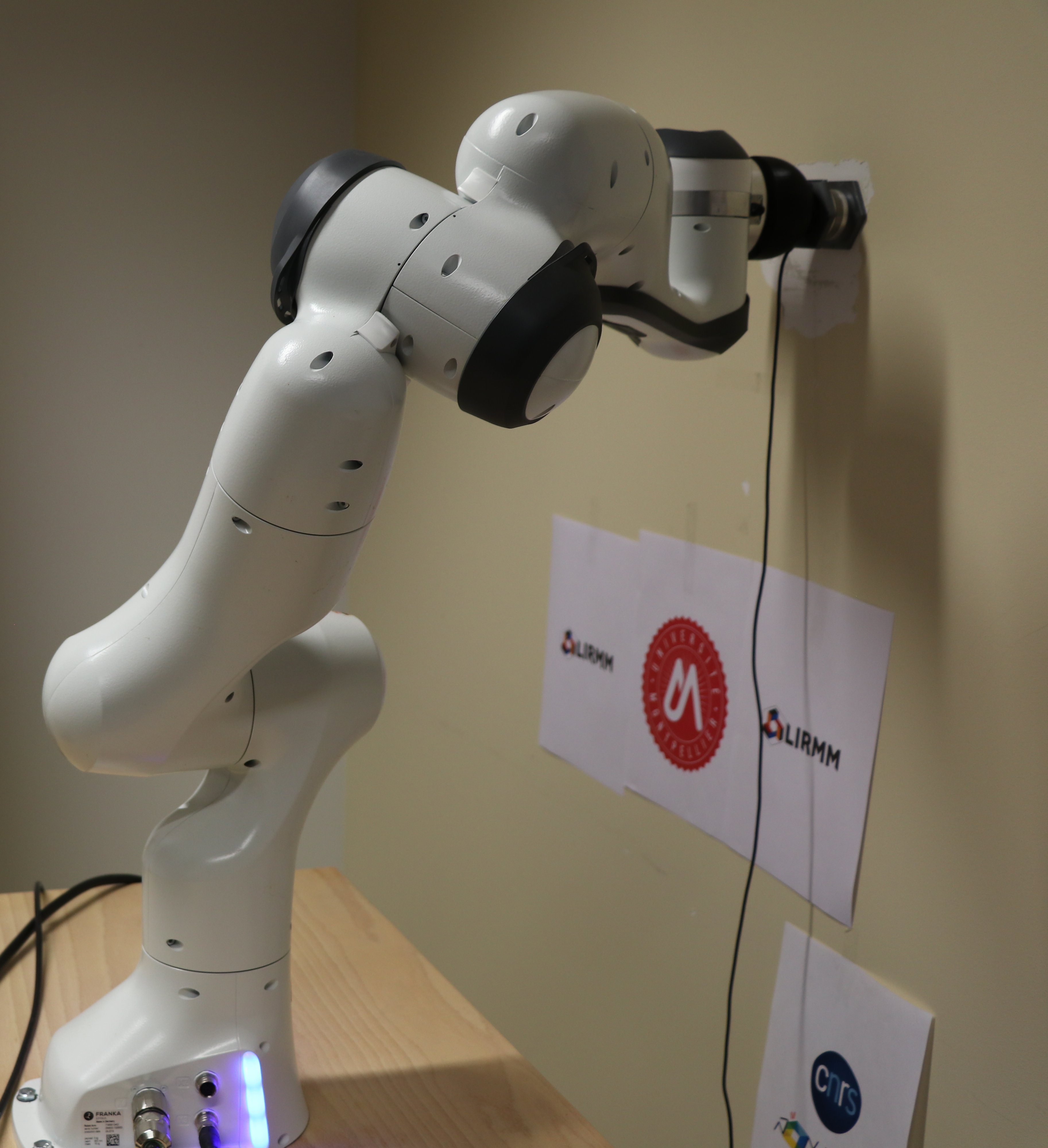}};
    \end{scope}
    \end{tikzpicture} 
    }
} & 
\subfigure [The measured $\jump \jvelocities$] {
    \resizebox{0.25\textwidth}{!}{%
    \begin{tikzpicture}
      \label{fig:measured_qd_jump_two}
    \begin{scope}
      \node[anchor=south west,inner sep=0] (image) at (0,0) {
        \includegraphics[width=0.24\textwidth]{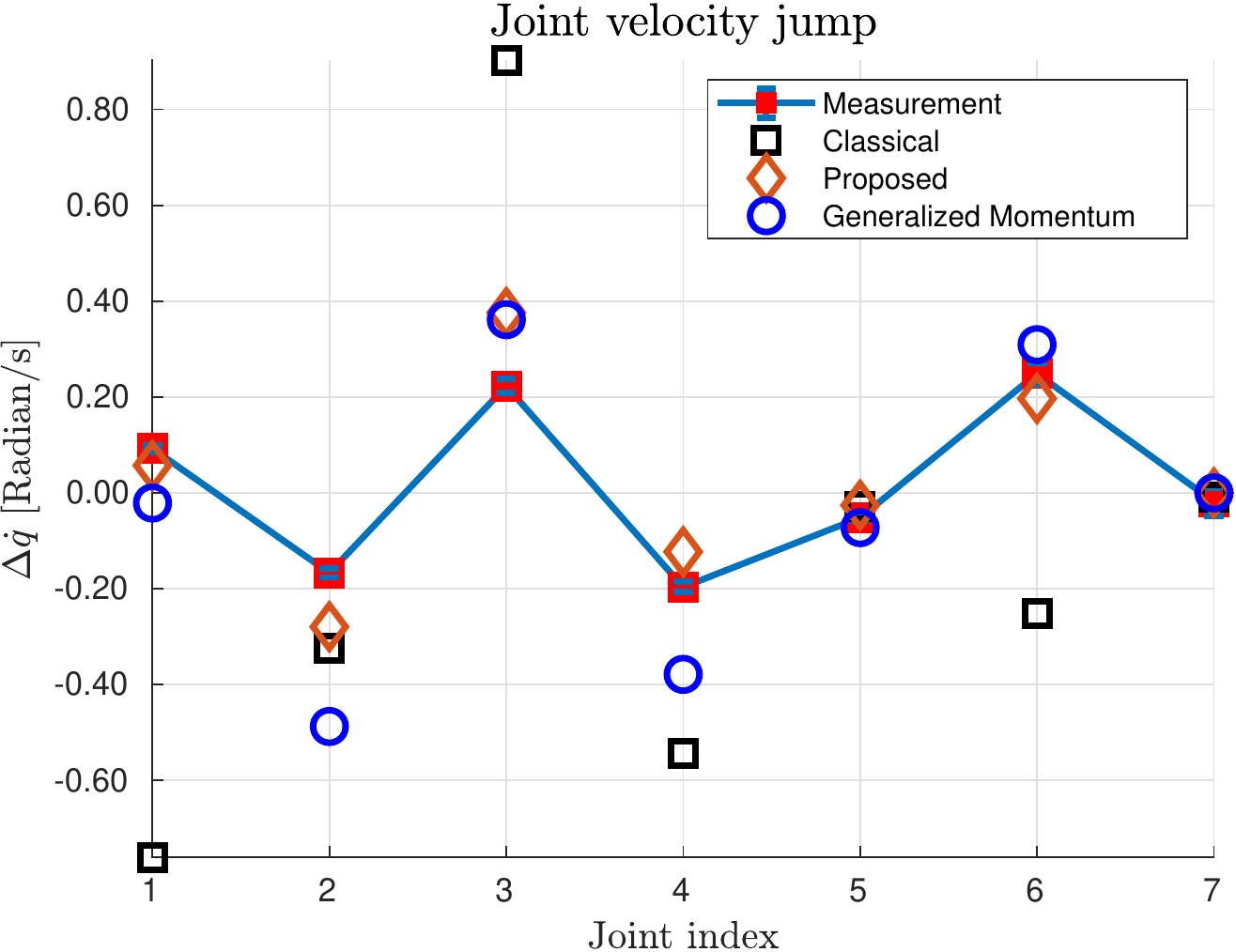}
      };
    \end{scope}
    \end{tikzpicture} 
    }
    } \\
\subfigure [Configuration three] {
  \resizebox{0.17\textwidth}{!}{%
  \begin{tikzpicture}
    \label{fig:contact_posture_three}
    \begin{scope}
      \vspace{10mm}
      \node[anchor=north east,inner sep=5] (image) at (0,0) {
          \includegraphics[width=0.21\textwidth]{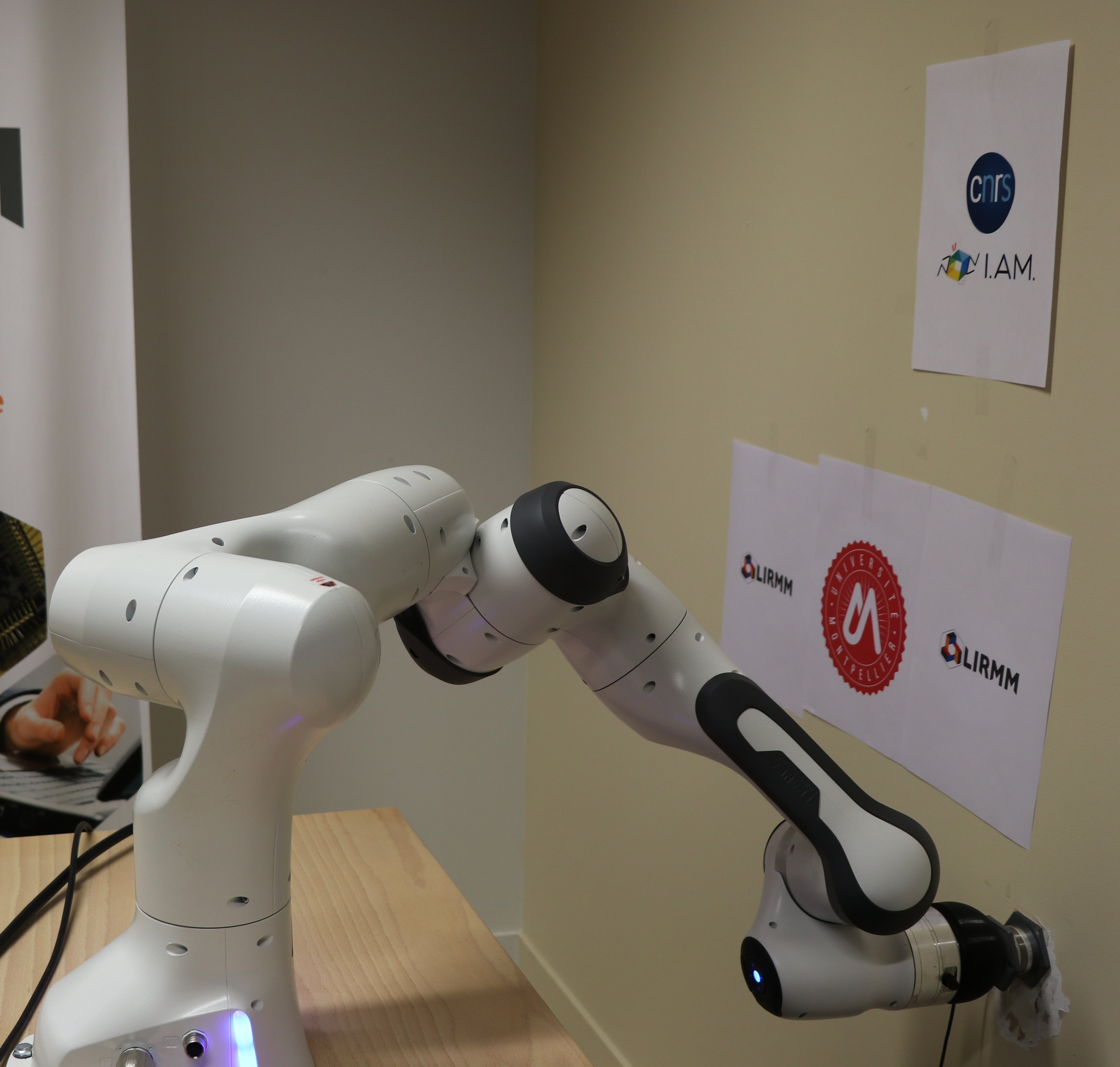}
        };
    \end{scope}
    \end{tikzpicture} 
    }
} & 
\subfigure [The measured $\jump \jvelocities$] {
    \resizebox{0.24\textwidth}{!}{%
    \begin{tikzpicture}
      \label{fig:measured_qd_jump_three}
    \begin{scope}
      \node[anchor=south west,inner sep=0] (image) at (0,0) {
        \includegraphics[width=0.24\textwidth]{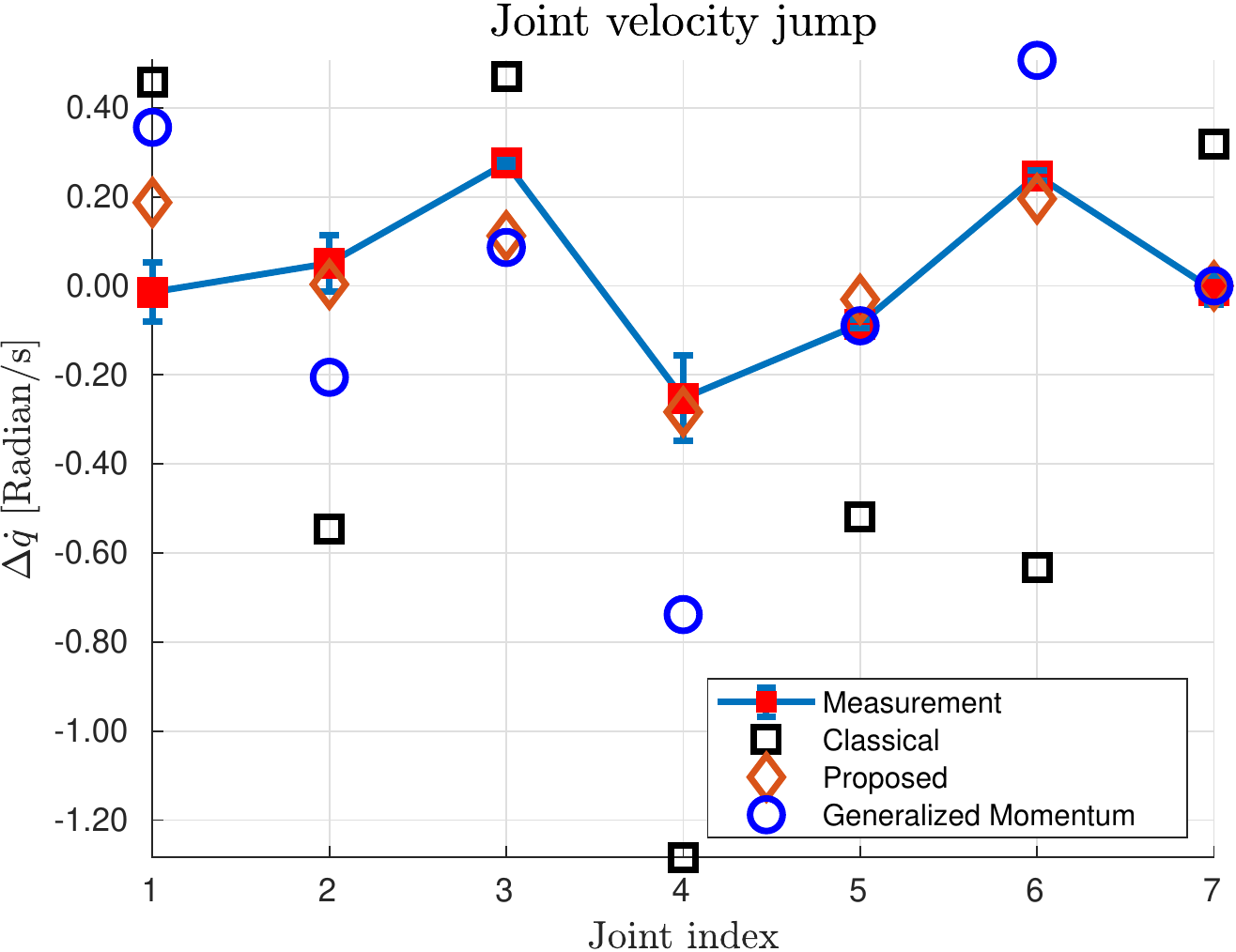}
      };
    \end{scope}
    \end{tikzpicture} 
    }
    }   
\end{tabular}
\caption{
  Comparison of different approaches when the reference-tangential-contact velocity is zero and  reference-normal-contact velocity is $0.18$\unitVelTS.
  We identify each $\jump \jvelocities_i$ in Fig.~\ref{fig:measured_qd_jump_one}, \ref{fig:measured_qd_jump_two}, and \ref{fig:measured_qd_jump_three} offline; see the experiment configurations in Sec.~\ref{sec:data}.
}
\label{fig:setup}
\vspace{-5mm}
\end{figure}
\subsection{The inaccurate impulse issue}
\label{sec:tImpulse_issue}
We showed in our earlier work~\cite[Fig.~9]{wang2022ral} that the impulse calculation~\eqref{eq:fl_impulse} is inaccurate as it overestimates the normal impulse~$\nImpulse$. Even more problematic, the calculation~\eqref{eq:fl_impulse} does not fulfill Coulomb’s friction cone since it does not take the friction coefficient $\coefF$ into account.

\begin{Example}
When the impact is frictionless $\coefF = 0$, the tangential impulse is zero~\cite[Sec.~2.4]{jia2017ijrr}. Given the joint positions $\jangles$ corresponding to the impact configurations in~\figRef{fig:contact_posture_one}, \ref{fig:contact_posture_two}, \ref{fig:contact_posture_three} and the coefficient of restitution $\coefR = 0$ and $\coefF = 0$, equation~\eqref{eq:fl_impulse} predicts:
\quickEq{eq:example_tangential_impulse}{
\impulse_1 = \vectorThree{
  0.4301}
{-0.2018}
{1.0993}
\quad \impulse_2 = \vectorThree{
  0.2783}
{0.4041}
{1.3776}
\quad \impulse_3 = \vectorThree{
  0.4952}
{-0.9224}
{2.1325}.
}
where the reference contact velocity expressed in the end-effector coordinates is: $[0, 0, 0.18]^\top$\unitVelTS.
It is obvious that the  tangential impulses in \eqref{eq:example_tangential_impulse} do not match $\coefF = 0$.
\end{Example}

\subsection{The ill-conditioning issue}
\label{sec:condition_number_issue}
Implementing the JSIM inversion in practice leads to critical numerical issues. The inversion amplifies errors of $\impulse$ in proportion to the high condition number of the JSIM; see the sources of numerical errors reported by Featherstone~\cite[Chapter~10.1]{featherstone2008book}, \cite[Sec.~2]{featherstone2004ijrr}. 
The high condition number of $\inertiaMatrix$ is an underlying property of the mechanism itself.
Inline with the analysis in~\cite{featherstone2004ijrr}, we obtain the following values  on the order of $\complexity{\dof^4}$ with $\dof =7$ 
for the impact configurations shown in \figRef{fig:contact_posture_one}, \ref{fig:contact_posture_two}, \ref{fig:contact_posture_three}:
\quickEq{eq:cd_M}{
2.9444 \times 10^3, \quad 3.1586 \times 10^3,~\text{and}\quad 3.9909 \times 10^3.
}
Similar, large condition numbers are reported for the KUKA LBR iiwa manipulator with seven joints \cite[Fig.~8]{fonseca2021task}.

\subsection{Unmodeled stiffness issue}
\label{sec:joint_motion_constraint}
We clarified in our earlier work that the impulse calculation is not accurate without considering stiffness behavior~\cite{wang2022ral}.
However, the classical approach~\eqref{eq:joint-space-solution} merely imposes the joint motion constraint
\quickEq{eq:kinematics}{
  \jacobian \jump \jvelocities = \jump\contactVel \hspace{0.3cm} \Leftrightarrow \hspace{0.3cm} \jacobian \jump \jvelocities - \jump\contactVel = \mathbf{0},
}
without modifying the task-space impulse-to-velocity mapping: 
\quickEq{eq:flexible_mapping}{
\jump \contactVel = {\jacobian \inverse{\inertiaMatrix} \transpose{\jacobian}}\impulse,
}
which is embeded in the intermediate step \eqref{eq:generalized_momentum}.
Given the same task-space impulse, \eqref{eq:flexible_mapping} predicts the same contact velocity jump for (1) a high-gain joint-position-controlled robot, and (2) a compliant pure joint-torque-controlled robot.

\section{Novel Prediction of Joint Velocity Jumps}
\label{sec:solution}
In response to the above issues, we (i) adopt the improved impulse calculation relying on the CRB assumption \cite{wang2022ral}, see \secRef{sec:impulse_calculatio};
(ii) avoid the JSIM inversion by solving $\jump \jvelocities$ from  momentum conservation equations in the task-space, see \secRef{sec:task-space};
and (iii) modify the task-space impulse-to-velocity mapping assuming
the robot is a composite-rigid body prior to the impact, see \secRef{sec:iim}.


\subsection{Impulse calculation}
\label{sec:impulse_calculatio}
We calculate the normal impulse $\nImpulse \in \RRv{}$ \cite[Eq.~32]{wang2022ral}~as:
\quickEq{eq:zImpulseCE}{
  \nImpulse = - (1 +\coefR) \frac{\preImpact{\nContactVel}}{\transpose{\bs{n}} \iim \bs{n}} \,\, ,
}
where $\bs{n}\in\RRv{3}$ is the surface normal at impact, $\iim \in \RRm{3}{3}$ is the widely applied IIM in the impact dynamics literature~\cite{stronge2000book,jia2017ijrr,khulief2013modeling,jia2019ijrr}. We derive two variants of IIM in \secRef{sec:iim}. 

\subsection{Momentum conservation in  the task-space}
\label{sec:task-space}
Given the equivalent inertia matrix  $\eInertiaMatrix \in \RRm{6}{6}$, the Newton-Euler's equations at the contact point are \cite[Eq.~4.16]{murray1994book}\cite[Eq.~2.68]{featherstone2008book}:
\quickEq{eq:NE_taskspace}{
  \eInertiaMatrix \vectorTwo{\dot{\contactVel}}{\dot{\rVel}} + 
  \dualCross{\vectorTwo{\contactVel}{\rVel}}{\eInertiaMatrix}\vectorTwo{\contactVel}{\rVel}
  = \vectorTwo{\force}{\moment},
}
where $\rVel\in\RRv{3}$ is the rotational velocity, $\moment\in\RRv{3}$ is the rotational moment,
and we adopt the six-dimensional cross product $\dualCross{}{}$ \cite{featherstone2008book}. 
Integrating \eqref{eq:NE_taskspace} over the impact duration $\impactDuration$, and neglecting the centrifugal and Coriolis forces concerning assumption \ref{assumption1}, yields
\quickEq{eq:id_contact}{
  \eInertiaMatrix  \vectorTwo{\jump \contactVel}{\jump \rVel} = \vectorTwo{\impulse}{0},
}
where $\int^{\impactDuration}_0\moment dt = 0$ (assumption~\ref{assumption3}).
Define $\iim$ as the $3\times 3$ corner matrix of $\inverse{\eInertiaMatrix}$, we 
left-multiply  $\inverse{\eInertiaMatrix}$ to \eqref{eq:id_contact}:
\quickEq{eq:task-space-momentum}{
  \jump \contactVel  = \iim \impulse.
}
Substituting the joint motion constraint \eqref{eq:kinematics} into \eqref{eq:task-space-momentum} yields
\quickEq{eq:task-space-momentum-conservation}{
\jacobian \jump \jvelocities = \iim \impulse.
}
Now, we obtain $\solution$ in a least-squares form 
\quickEq{eq:task-space-solution}{
  \solution = \argmin_{\jump \jvelocities} \|\jacobian \jump \jvelocities - \iim \impulse \|^2
  = \pseudoInverseRowDef{\jacobian} \iim \impulse,
}
which does not rely on the critical JSIM inversion. 

\subsection{Imposing the composite-rigid-body assumption}
\label{sec:iim}
We compute the equivalent inertia matrix $\eInertiaMatrix$ at the contact point based on the CRB assumption.
Defining the centroidal frame $\cframe{\com}$ between the inertial frame $\cframe{\inertialFrame}$ and the contact frame $\cframe{\contactPoint}$,  the \emph{body velocity}\footnote{$\bodyVel{\inertialFrame}{i}$ is noted by $\vc{V}^b_{{\inertialFrame}{i}}$ in \cite{murray1994book}. We omit the subscript $^b$ since we only use body velocities. } $\bodyVel{\inertialFrame}{\contactPoint} \in \RRv{6}$  associated with the rigid motion
$ g_{{\inertialFrame}{\contactPoint}}(t) \in SE(3)$ writes \cite[Proposition~2.15]{murray1994book}:
\quickEq{eq:contact_vel_transform}{
  \bodyVel{\inertialFrame}{\contactPoint} = \bodyVelTransform{\com}{\contactPoint}{\inertialFrame},
}
where $\bodyVel{\com}{\contactPoint}$ is the relative velocity \cite{orin2013auro}. According to the CRB  assumption, we impose zero relative velocity $\bodyVel{\com}{\contactPoint} = 0$. Thus, we approximate:
\quickEq{eq:jump_vel_approximation}{
\jump \bodyVel{\inertialFrame}{\contactPoint}
\approx
\twistTransform{\com}{\contactPoint}\jump \bodyVel{\inertialFrame}{\com}
}
where $\twistTransform{\com}{\contactPoint}$ transforms $\bodyVel{\inertialFrame}{\com}$ from  $\cframe{\com}$ to $\cframe{\contactPoint}$.

Given the centroidal inertia \cite{orin2013auro} $\crbGInertia \in \RRm{6}{6}$, the centroidal momentum is:
$$
\crbGInertia \jump \bodyVel{\inertialFrame}{\com} =  \geometricFT{\contactPoint}{\com}\vectorTwo{\impulse}{0},
$$
where $\geometricFT{\contactPoint}{\com} \in \RRm{6}{6}$ transforms $\vectorTwo{\impulse}{0}$ from $\cframe{\contactPoint}$ to $\cframe{\com}$.
Substituting  $\jump \bodyVel{\inertialFrame}{\com} =  \inverse{\crbGInertia}\geometricFT{\contactPoint}{\com}\vectorTwo{\impulse}{0}$ into \eqref{eq:jump_vel_approximation}, we have
$$
\jump \bodyVel{\inertialFrame}{\contactPoint}
 \approx  \twistTransform{\com}{\contactPoint} \inverse{\crbGInertia}\geometricFT{\contactPoint}{\com}\vectorTwo{\impulse}{0}.
$$

The linear part of  $\bodyVel{\inertialFrame}{\contactPoint}$ is the contact velocity $\contactVel$ in its body coordinates\footnote{Note that representing a quantity in a far-away coordinate frame introduces another source of numerical error \cite[Chapter~10.1]{featherstone2008book}.} $\bodyTV{\inertialFrame}{\contactPoint} \in \RRv{3}$.
Thus, we extract the linear part (the upper-left $3\times 3$ matrix) of
$$
\begin{aligned}
\quad & \twistTransform{\com}{\contactPoint} \inverse{\crbGInertia} \geometricFT{\contactPoint}{\com} \\
&= \twistTransformDef{\com}{\contactPoint}
\matrixTwo{\frac{1}{\mass}\identityMatrix_{3\times 3}}{0}{0}{\inverse{\metaInertia}}
 \geometricFTDef{\contactPoint}{\com}
\end{aligned}
 $$
 to have the IIM ${\iimcrb} \in \RRm{3}{3}$:
\quickEq{eq:iim_inertia_def}{
  \jump\bodyTV{\inertialFrame}{\contactPoint}  =
  \underbrace{(\frac{1}{\mass}\identityMatrix_{3\times 3} - \rotationInv{\com}{\contactPoint}\translationSkew{\com}{\contactPoint} \inverse{\metaInertia} \translationSkew{\com}{\contactPoint}\rotation{\com}{\contactPoint} )}_{\iimcrb}
  \impulse,
}
where $\metaInertia \in \RRm{3}{3}$ is the CRB moment of inertia, and $\rotation{\com}{\contactPoint} \in \RRm{3}{3}, \translation{\com}{\contactPoint} \in \RRv{3}$ are the rotation and translation respectively, between the COM and the contact point. The skew-symmetric matrix $\translationSkew{\com}{\contactPoint} \in \RRm{3}{3}$ represents the cross product as matrix multiplication.

We calculate $\crbImpulse$ by substituting $\iim = \iimcrb$ into \eqref{eq:zImpulseCE}.
Then we predict $\proposedSolution$ by substituting both  $\impulse = \crbImpulse$ and   $\iim = \iimcrb$ into \eqref{eq:task-space-solution}:
\quickEq{eq:proposed_projection}{
  \proposedSolution= \argmin_{\jump \jvelocities} \|\jacobian \jump \jvelocities - \iimcrb \crbImpulse \|^2
  = \pseudoInverseRowDef{\jacobian} \iimcrb \crbImpulse.
}
%
%

Other models relying on joint-space (generalized) momentum conservation \cite{lankarani2000poisson,khulief2013modeling}
derive the IIM without the CRB assumption: 
\quickEq{eq:iim_gm}{
  \jacobian \jump \jvelocities = \underbrace{\jacobian \inverse{\inertiaMatrix} \transpose{\jacobian}}_{\iimgm} \impulse,
}
which happens to be the same as the intermediate step \eqref{eq:generalized_momentum} of the classical approach.

For comparison, we substitute $\iim = \iimgm$ into \eqref{eq:zImpulseCE} for $\gmImpulse$. Note that $\gmImpulse$ is different  from $\classicalImpulse$, see  \cite[Fig.~9]{wang2022ral}. 
Then, we substitute $\iim = \iimgm$ and $\impulse = \gmImpulse$ into \eqref{eq:task-space-solution} to obtain another prediction:
\quickEq{eq:gm_projection}{
  \gmSolution = 
\argmin_{\jump \jvelocities} \|\jacobian \jump \jvelocities - \iimgm \gmImpulse \|^2
= \pseudoInverseRowDef{\jacobian} \iimgm \gmImpulse.
}
We refer to $\gmSolution$ as the generalized-momentum solution.

\section{Data Acquisition}
\label{sec:data}

We collected a large amount of impact data with the FrankaEmika Panda robot
in order to thoroughly evaluate and compare the three predictions  $\classicalSolution, \proposedSolution, \gmSolution$.
Specifying the contact velocities in the inertial frame $\cframe{\inertialFrame}$, 
we conducted two types of impact experiments:
\begin{enumerate}
\item Commanding $\xContactVel = \yContactVel = 0$\unitVelTS, the robot end-effector impacted  with three different joint postures shown in \figRef{fig:contact_posture_one}, \ref{fig:contact_posture_two}, \ref{fig:contact_posture_three} and 5 reference normal contact velocities: $\nContactVel = 0.08, 0.10, 0.12, 0.15, 0.18$\unitVelTS.
\item Commanding the normal contact velocity  $\nContactVel = 0.12$\unitVelTS,
  the robot applied an increasing tangential velocity $\yContactVel = 0.03$, $0.04$, $0.05$, $0.06$, $0.07$, $0.08$, $0.09$, $0.10$, $0.11$, $0.12$\unitVelTS with the end-effector
  while keeping $\xContactVel = 0$\unitVelTS. When the impact took place, the joint postures in all the situations were close to \figRef{fig:contact_posture_one}.
\end{enumerate}
Repeating each configuration ten times, 
we collected 150 experiments in the first case and another 100 experiments in the second case\footnote{We presented the contact-force profiles of the 150 experiments in our earlier work \cite{wang2022ral} for normal impulse calculation.  In this paper, we present the joint velocity jumps $\jump\jvelocities$ collected from the same 150 experiments. The data of the other 100 experiments have been specifically collected for this article.}.

We compute $\classicalSolution, \proposedSolution, \gmSolution$ utilizing the mean measurement $\measured{\jangles}$, $\measured{\jvelocities}$ of the ten repetitions in each configuration\footnote{We apply the Panda robot model documented at \url{https://github.com/jrl-umi3218/mc_panda}.}, and compared to the corresponding mean measurement~$\jvelocitiesJumpEstimate$.
The $\measured{\jangles}, \measured{\jvelocities}$, and $\jvelocitiesJumpEstimate$  were consistent within the repetitions. 
The standard deviation of $\measured{\jangles}$, $\measured{\jvelocities}$  are $0.000954$\unitPosJS and $0.0042$\unitVelJS, respectively.
The standard deviation of measured $\jvelocitiesJumpEstimate$ is higher, yet still small, see \figRef{fig:qd_jump_std}.


We measured the ground-truth impulse with an ATI-mini45 force-torque sensor.  It can capture the impact dynamics with a sampling rate of 25000\unitHz without low-pass filtering.
We apply the average measured impulse from ten identical experiments to remove the random effects.

The friction coefficient for all experiments is $\coefF = 0.1123$. The impulse $\impulse$ fulfills Coulomb's friction cone~\cite[Eq.~4.8]{stronge2000book,jia2017ijrr}:
  \quickEq{eq:columb_friction_cone}{
    \sqrt{\impulse^2_x + \impulse^2_y } \leq \coefF \impulse_z.
  }

Regardless of the joint configurations or contact velocities, the impact duration was about
15\unitMs on average, and
25\unitMs at maximum for the Panda robot \cite[Fig.~6]{wang2022ral}. The control cycle was 1\unitMs. 
Upon contact detection, the controller immediately pulled back the end-effector along the contact normal.

\subsection{Measuring the joint velocity jumps}
\label{sec:measure_qd_jump}
\begin{figure}[hbtp]
  \vspace{-3mm}
  \centering
  \includegraphics[width=\columnwidth]{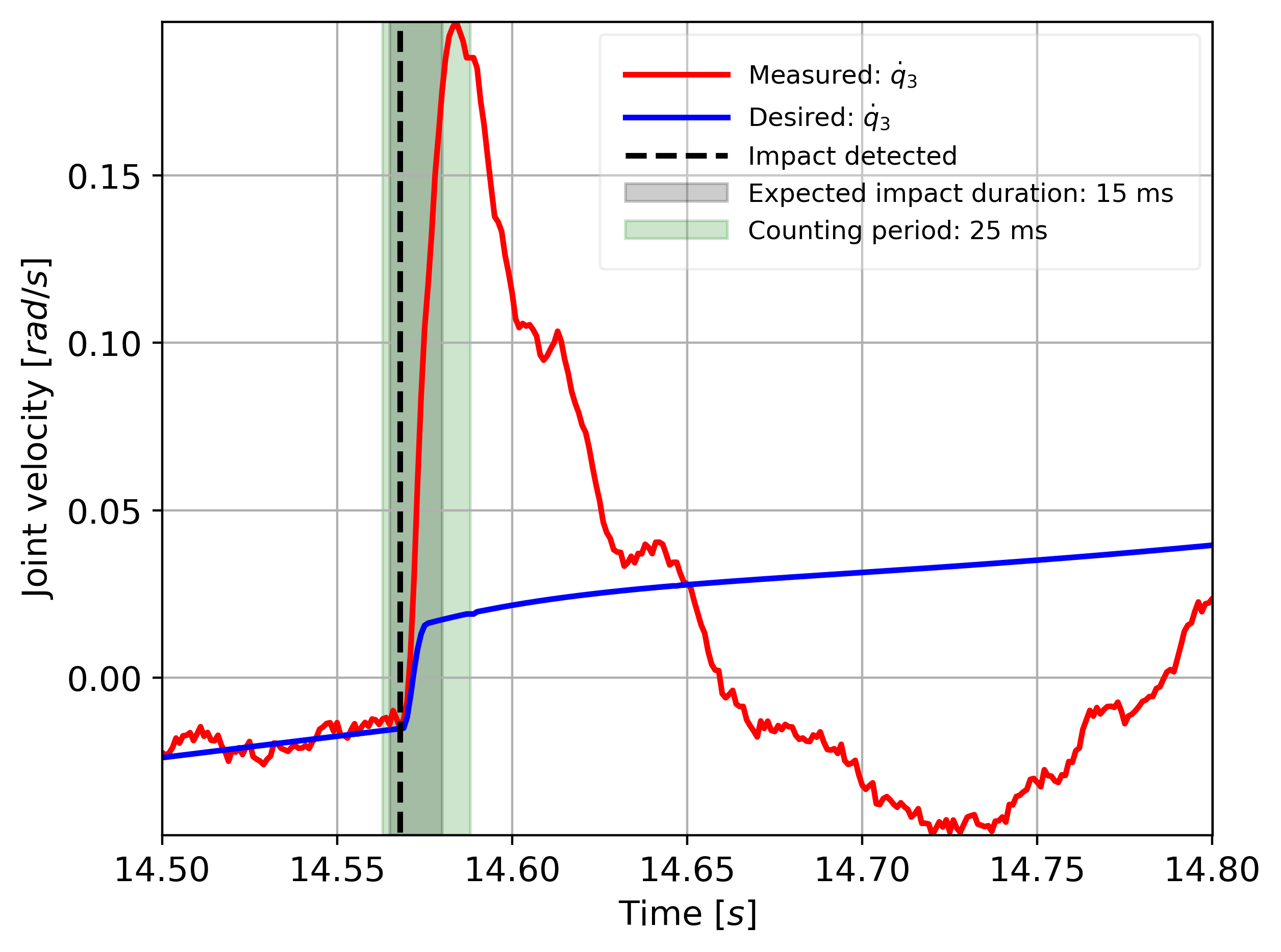}
  \caption{For impact configuration in Fig.~\ref{fig:contact_posture_one}, we plot the measured (red-solid) and desired joint velocities (blue-solid) before and after the impact detection (black-dashed). Please note the correspondence with joint 3 in Fig.~\ref{fig:measured_qd_jump_one}. The prior knowledge of the impact duration $\impactDuration$ is important. Otherwise, we could find a different $\jvelocitiesJumpEstimate$ in a larger time interval. Due to the space limit, we document the same plot for the other joints online: \url{https://gite.lirmm.fr/yuquan/fidynamics/-/wikis/Impact-Example-One}.
}
  \label{fig:dq_examples}
  \vspace{-3mm}
\end{figure}

The impact-detection time is $3$ to $5$\unitMs by thresholding the joint torque error~\cite[Sec.~VI]{wang2022ral}.
In order to measure the joint velocity jumps~$\jvelocitiesJumpEstimate$, 
we search for in a conservative time window, covering 5\unitMs before and 20\unitMs  after the impact detection; see \figRef{fig:dq_examples}.
\begin{figure*}[hbtp]
  \centering
  \includegraphics[width=0.72\textwidth]{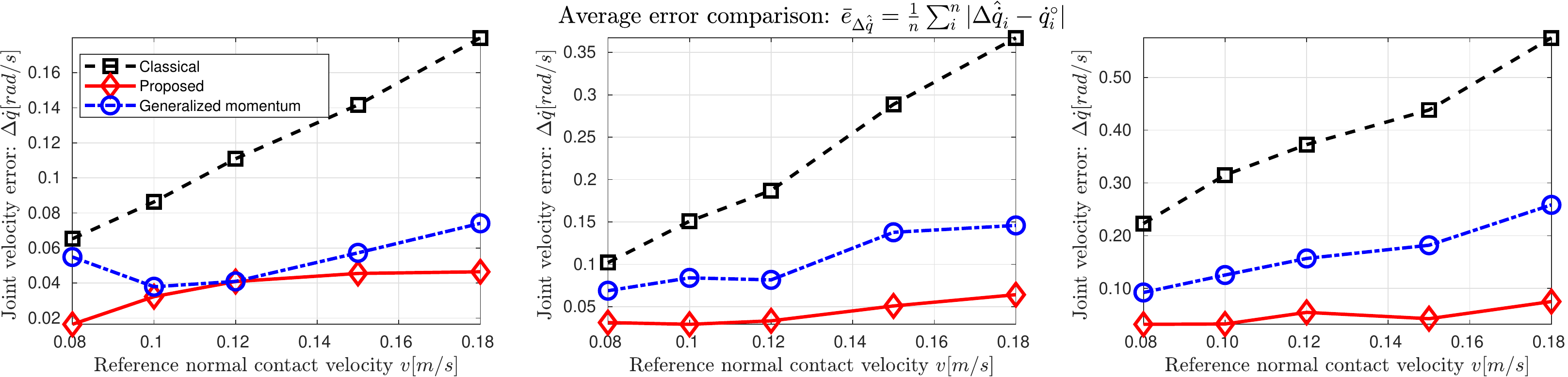}
  \includegraphics[width=0.21\textwidth]{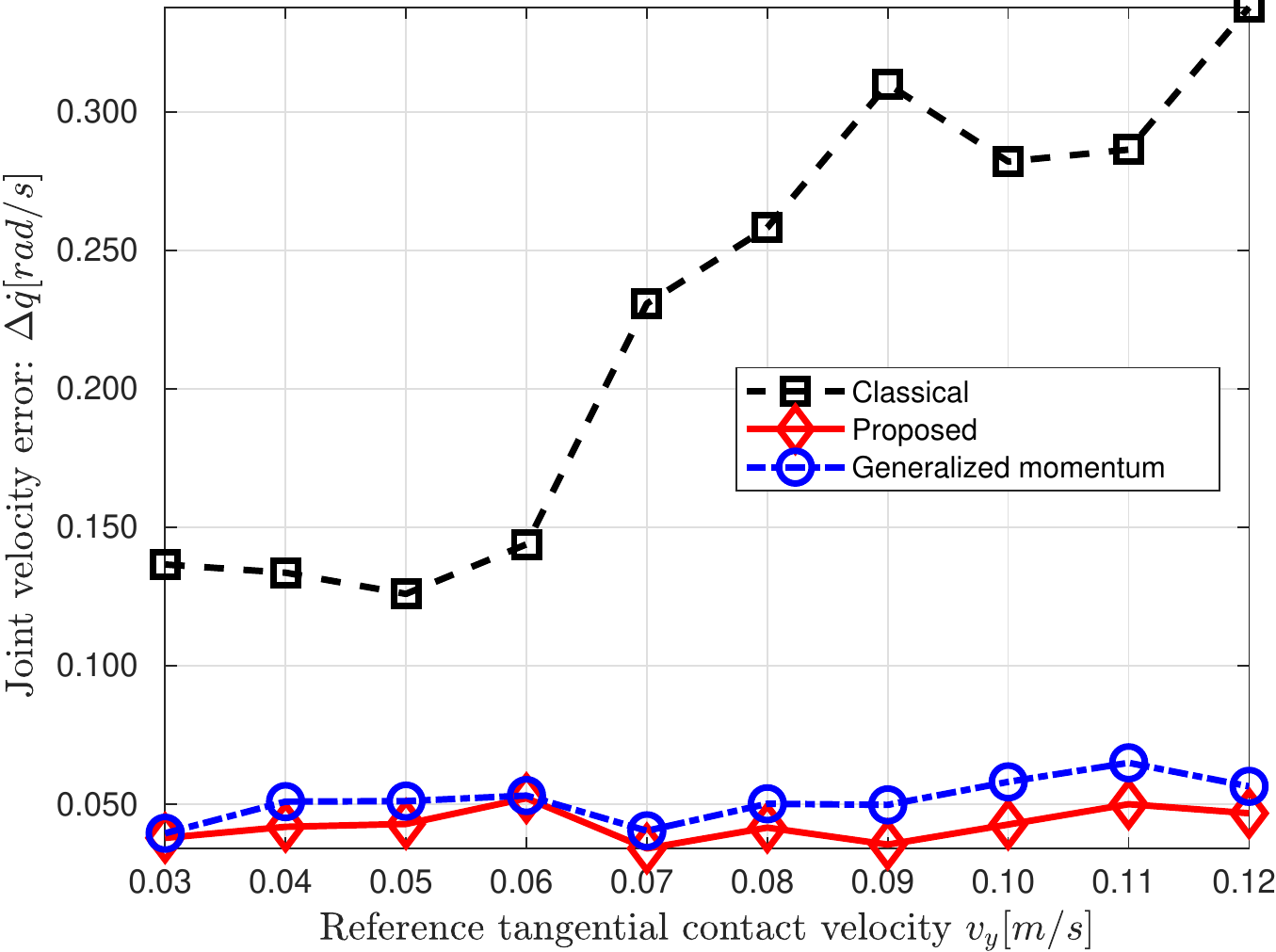}
  \caption{
    From left to right, the average prediction error \eqref{eq:relative_error} of  the impact configurations in Fig.~\ref{fig:contact_posture_one}, \ref{fig:contact_posture_two}, \ref{fig:contact_posture_three}, and the last one corresponds to the second set of experiments with non-zero tangential velocities.
}
\vspace{-3mm}
\label{fig:average_error}
\end{figure*}

We plot the mean of the measured $\jvelocitiesJumpEstimate$ in \figRef{fig:qd_jump_mean}.
Given the increasing reference normal contact velocities $0.08, 0.10, 0.12, 0.15, 0.18$\unitVelTS, 
the magnitude of $\jvelocitiesJumpEstimate_i$ for $i = 1\ldots, \dof$ increases  without changing the sign (jump direction).


\subsection{Prediction error definitions}
\label{sec:error}
We define the \emph{absolute prediction error} $\error{ \jump \jvelocities} \in \RRv{7}$ as:
\quickEq{eq:error_abs}{
\error{ \jump \jvelocities^*} =  \jump \jvelocities^* - \jvelocitiesJumpEstimate,
}
We plot  the predictions:  $\classicalSolution$ \eqref{eq:textbook_dq_jump}, $\proposedSolution$ \eqref{eq:proposed_projection},   and  $\gmSolution$ \eqref{eq:gm_projection}, 
in \figRef{fig:qd_jump_error_textbook}, \figRef{fig:qd_jump_error_proposed}, and \figRef{fig:qd_jump_error_gm}, respectively.
To compare the different options by a scalar measure, 
we define the \emph{average prediction error} $\aError{\jump\jvelocities} \in \RRv{}$:   
\quickEq{eq:relative_error}{
  \aError{\jump \hat{\dot{\bs{q}}}} = 
  \frac{1}{\dof}\agg{i}{\dof}{
    \abs{\jump \hat{\dot{\bs{q}}}_i - \jvelocitiesJumpEstimate_i}}.
}

\section{Comparison results and discussion}
\label{sec:experiments}

The first set of experiments reveal that:
\begin{itemize}
\item Given the \emph{average prediction error}  \eqref{eq:relative_error},
  our proposed prediction $\proposedSolution$ outperforms $\classicalSolution$ by 81.98$\%$.
  We list the numeric values in \tableRef{tab:average_error} and plot them in \figRef{fig:average_error}.
  We also plot the \emph{absolute prediction error} \eqref{eq:error_abs} of each joint in \figRef{fig:qd_jump_error_textbook}, \figRef{fig:qd_jump_error_proposed}, and \figRef{fig:qd_jump_error_gm}, respectively.
\item $\proposedSolution$ only requires analytical computation. It is robust to impulse-calculation errors,
  \secRef{sec:robustness}.
\end{itemize}

The second set of experiments reveal that: 
\begin{itemize}
\item The proposed prediction $\proposedSolution$ applies to non-zero tangential velocities, see \secRef{sec:non-zero-preImpact-tangential-vel}.
\end{itemize}

We apply the impulse calculation \eqref{eq:zImpulseCE} for $\proposedSolution$ and $\gmSolution$ with $\iim = \iimcrb$ and $\iim = \iimgm$, respectively.
We ignore the tangential component of $\classicalImpulse$ for $\classicalSolution$ due to the discussion  in \secRef{sec:tImpulse_issue}.
The contact mode was sliding in all the experiments due to the discussion in \secRef{sec:non-zero-tangential-vel-reason}.

\begin{table}[htbp!]
  \caption{
    Average prediction error  \eqref{eq:relative_error} of
    $\proposedSolution$, $\classicalSolution$, and $\gmSolution$.
    All the values share the same unit: \unitVelJS, and are plotted in Fig.~\ref{fig:average_error}.
  }
\begin{tabular}{l*{4}{c}r}
  Reference:\hspace{-3.2mm} & $0.08$~m/s\hspace{-3.2mm} & $0.10$~m/s\hspace{-3.2mm} & $0.12$~m/s\hspace{-3.2mm} & $0.15$~m/s\hspace{-3.2mm} & $0.18$~m/s\hspace{-3.2mm}    \\
  \hline
  \multicolumn{6}{c}{The proposed solution average error $\aError{\proposedSolution}$ \eqref{eq:proposed_projection}}\\
  \hline
  Fig.~\ref{fig:contact_posture_one}  &$0.0165$ & $0.0323$ & $0.0409$ & $0.0455$ &$0.0464$ \\ 
  Fig.~\ref{fig:contact_posture_two} &$0.0313$  &$0.0293$ &$0.0334$   &$0.0509$  &$0.0642$\\
  Fig.~\ref{fig:contact_posture_three} &$0.0321$  &$0.0327$  & $0.0547$  & $0.0428$  &$0.0752$ \\
  \hline
  \multicolumn{6}{c}{The classical solution average error $\aError{\classicalSolution}$ \eqref{eq:textbook_dq_jump}}\\
  \hline
  Fig.~\ref{fig:contact_posture_one}  &$0.0653$ &$0.0863$  & $0.1109$  &$0.1417$  & $0.1799$\\  
  Fig.~\ref{fig:contact_posture_two}  &$0.1020$  &$0.1509$  &$0.1871$  &$0.2887$  &$0.3672$\\
  Fig.~\ref{fig:contact_posture_three} &$0.2231$  &$0.3149$  &$0.3726$  &$0.4382$  &$0.5748$\\
  \hline
  \multicolumn{6}{c}{The generalized-momentum average solution $\aError{\gmSolution}$ \eqref{eq:gm_projection}}\\
  \hline
  Fig.~\ref{fig:contact_posture_one} &$0.0549$  &$0.0379$  &$0.0409$  &$0.0572$  &$0.0741$\\
  Fig.~\ref{fig:contact_posture_two} &$0.0689$  &$0.0840$  &$0.0817$  &$0.1378$  &$0.1459$\\
  Fig.~\ref{fig:contact_posture_three} &$0.0924$  &$0.1255$  &$0.1569$  &$0.1820$  &$0.2584$\\
\end{tabular}
\label{tab:average_error}
\vspace{-3mm}
\end{table}

\subsection{Robustness to impulse-calculation errors}
\label{sec:robustness}
We denote the measured impulse as $\impulseEstimate$. 
According to our former benchmark experiments \cite[Fig.~9]{wang2022ral}, $\crbImpulse$ is most-close-to~$\measuredImpulse$. 
Further, $\gmImpulse$ is less than  the measured impulse $\measuredImpulse$  while  $\classicalImpulse$ overestimates.
Concerning the impulse calculation errors,
we substituted the measured impulse $\measuredImpulse$ (instead of the predicted impulse) into the different approaches for comparison. We denote  the predictions after the substitution as:
$\classicalSolution(\measuredImpulse)$ \eqref{eq:textbook_dq_jump}, $\gmSolution(\measuredImpulse)$ \eqref{eq:gm_projection}, and $\proposedSolution(\measuredImpulse)$ \eqref{eq:proposed_projection}.

The substitution introduced  a considerable impulse correction for the classical prediction $\classicalSolution$ and  the solution without considering high-stiffness behaviours $\gmSolution$. 

\begin{figure}[hbtp]
  \vspace{-2mm}
  \includegraphics[height=3.4cm]{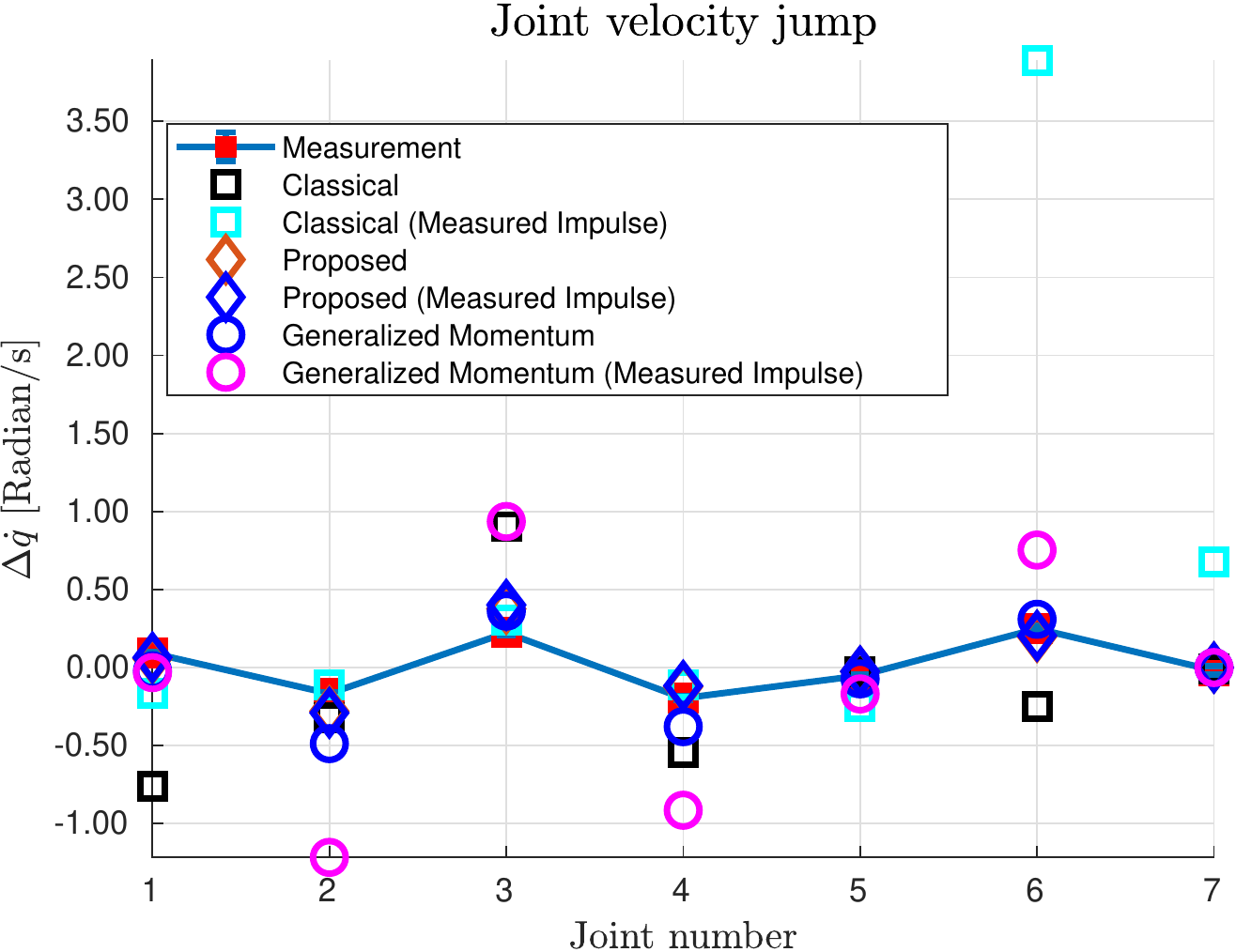}
  \includegraphics[height=3.4cm]{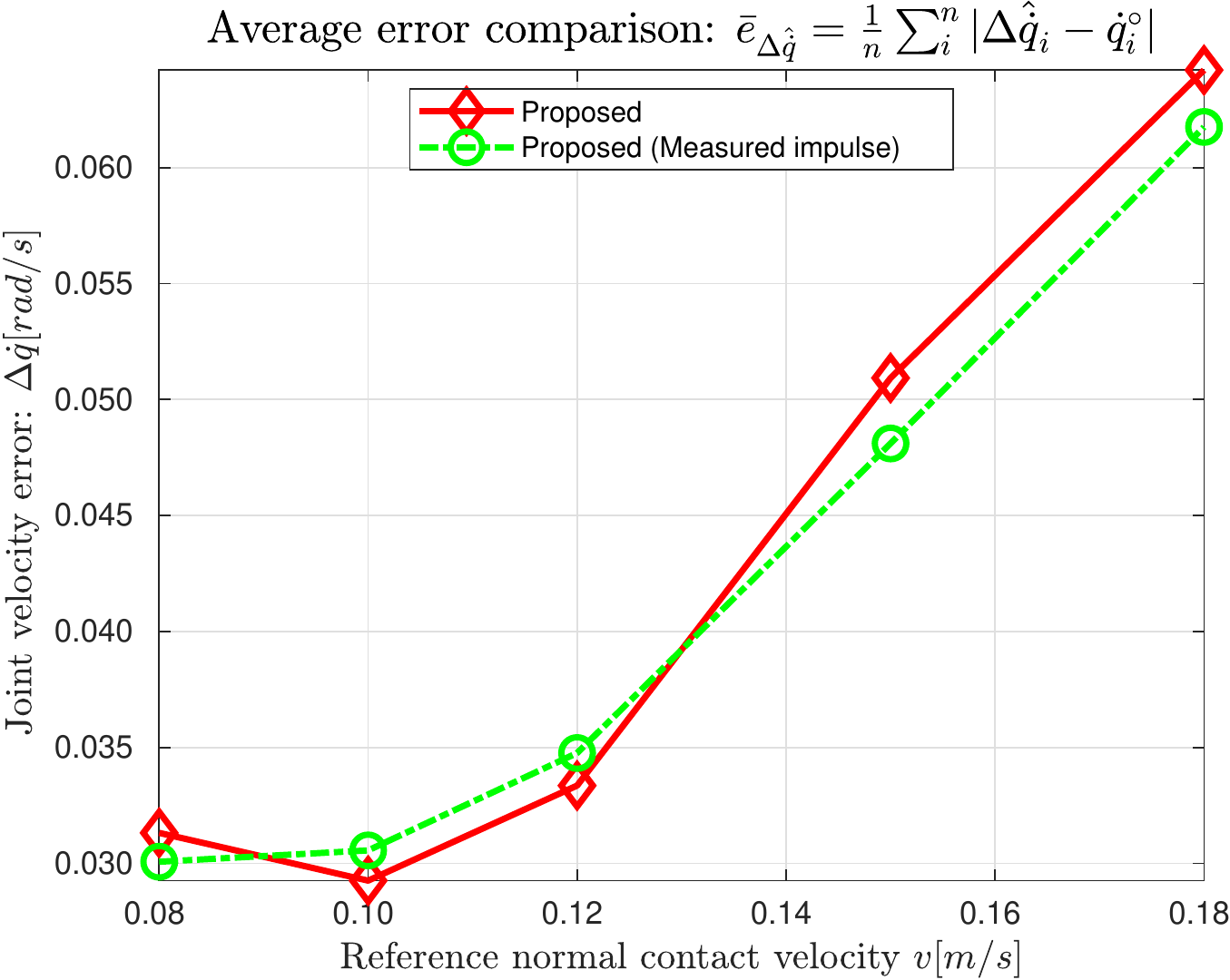}
  \caption{
    We compare  $\jump \jvelocities$ computed with the measured or calculated impulse for impact configuration two in Fig.~\ref{fig:contact_posture_two}. 
    {\bf Left:}  When the reference contact velocity is 0.18\unitVelTS, we plot the predicted $\jump \jvelocities$ by each joint.
    {\bf Right:}
    The  average error of $\proposedSolution(\measuredImpulse)$  remains on the same level regardless of the impulse calculation error.
  }
  \label{fig:robustness_issue}
  \vspace{-3mm}
\end{figure}
 
Given the errors on the left side of \figRef{fig:robustness_issue}, the accuracy of  $\classicalSolution(\measuredImpulse)$ or $\gmSolution(\measuredImpulse)$  decreases. Compared to  $\gmSolution(\measuredImpulse)$, the performance decrease of $\classicalSolution(\measuredImpulse)$ is even worse due to the high condition number of the JSIM.

$\proposedSolution(\measuredImpulse)$ keeps a similar average error concerning the right side of  \figRef{fig:robustness_issue}.
Since the only difference between computing $\proposedSolution$ and $\gmSolution$ is the choice of $\iim = \iimcrb$ or $\iim = \iimgm$, we conclude $\proposedSolution$ outperforms because of the CRB assumption.

\begin{figure}[hbtp]
  \vspace{-2mm}
  \centering
  \includegraphics[width=\columnwidth]{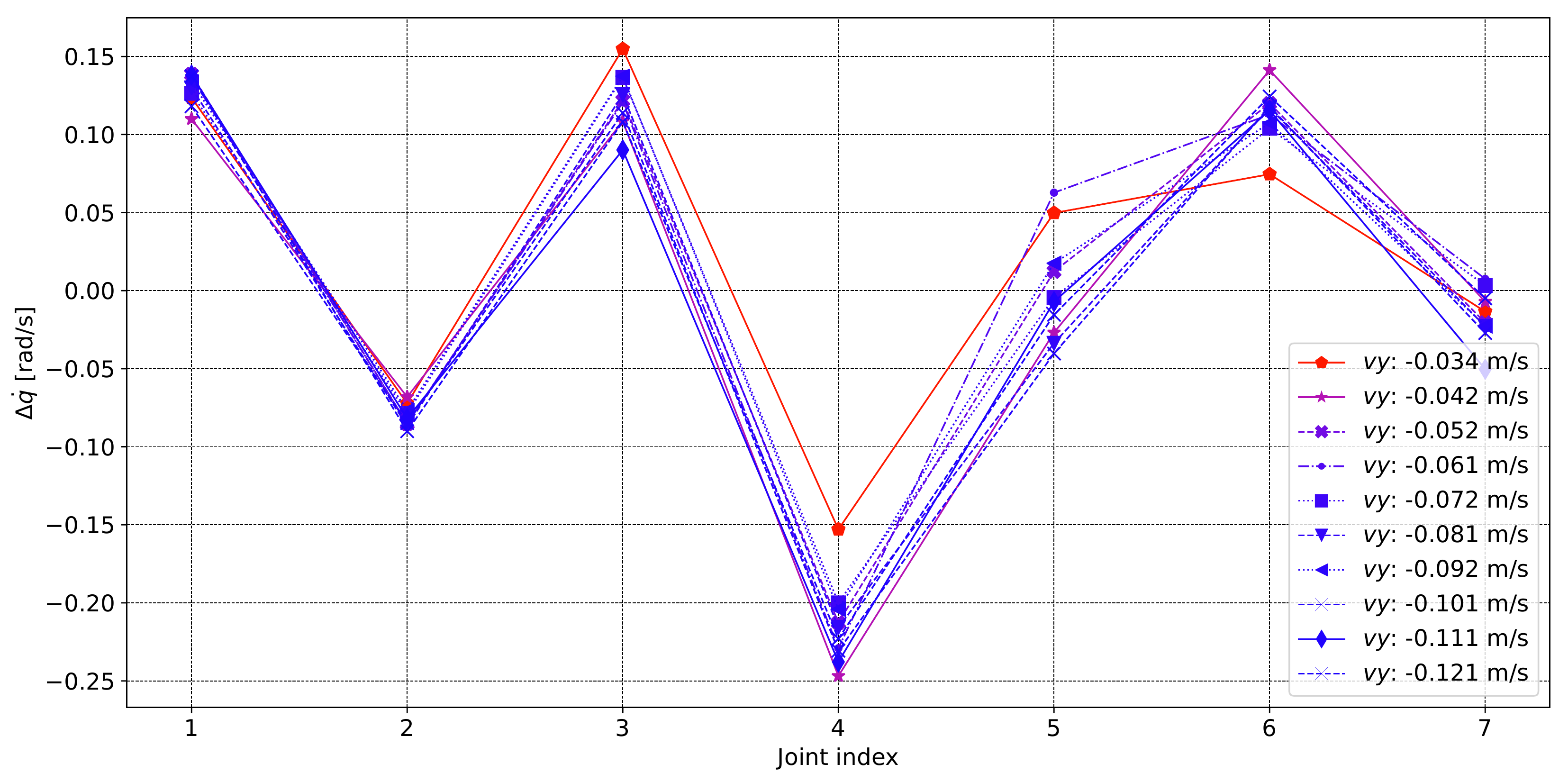}
  \caption{We plot the measured $\measured{\jump \jvelocities}$ by joint, given the reference $\nContactVel = 0.12$\unitVelTS,  $\xContactVel = 0$\unitVelTS,  and $\yContactVel$ increasing from $0.03$\unitVelTS to $0.12$\unitVelTS. Note that there is a discrepency between the measured  $\yContactVel$ and its reference. 
  }
  \label{fig:tangentialVel}
  \vspace{-3mm}
\end{figure}
\begin{figure}[hbtp]
  \centering
  \includegraphics[height=4.2cm]{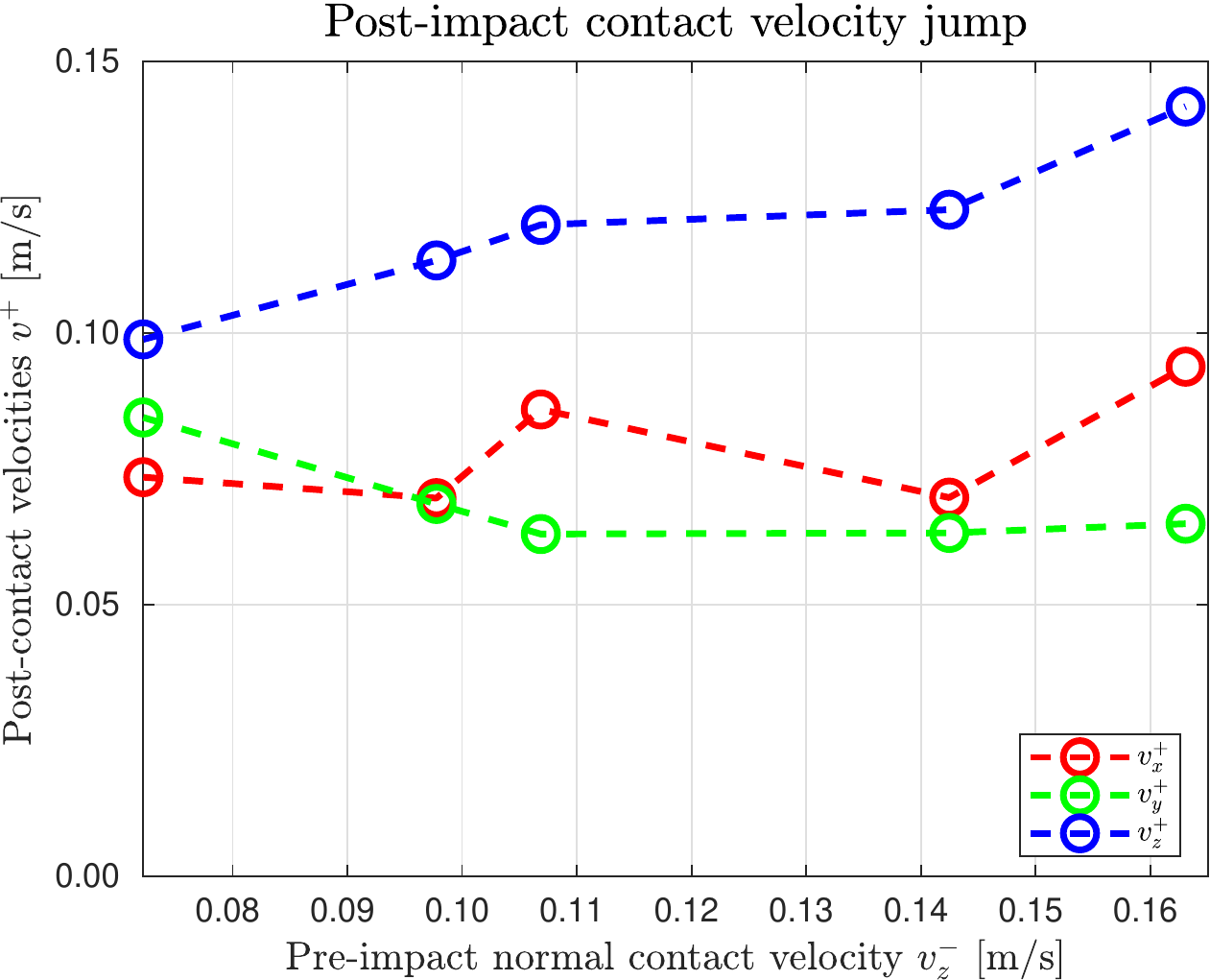}
  \includegraphics[height=4.2cm]{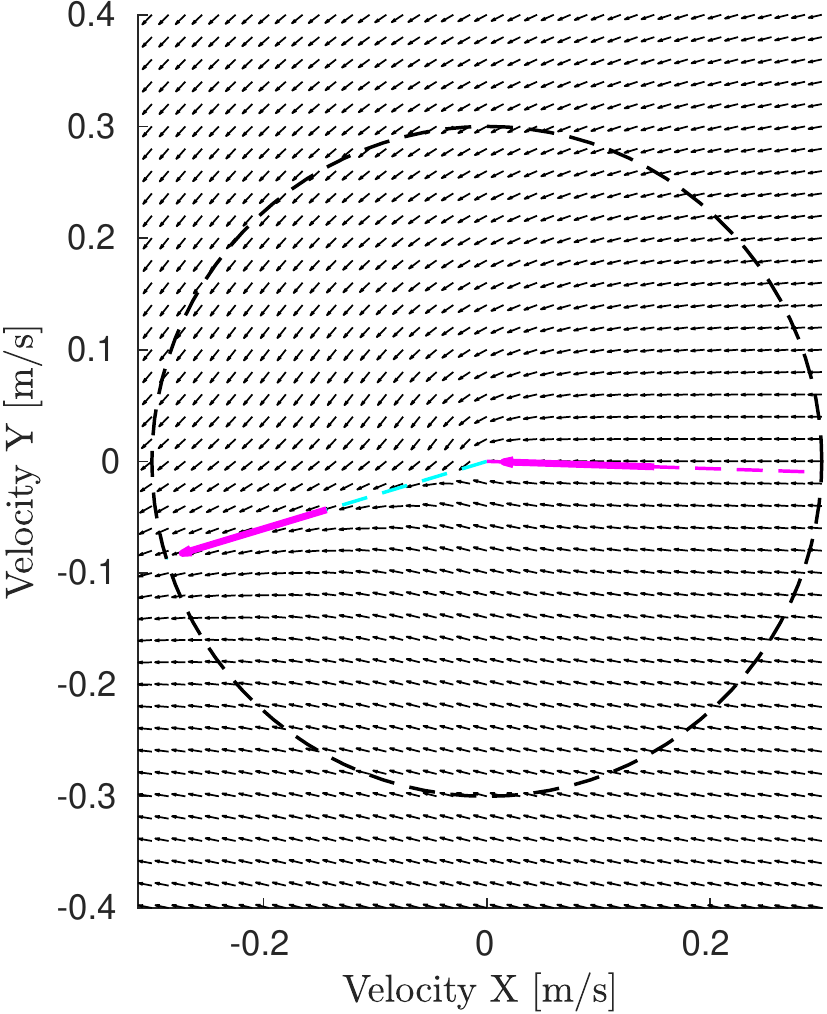}
  \caption{
    {\bf Left:}  The mean post-impact velocity $\postImpact{\contactVel}$ for impact configuration two in Fig.~\ref{fig:contact_posture_two}.
    {\bf Right:} The vector field of the tangential velocity $\tp{\contactVel}$ during the impact event. We marked the two invariant directions by pink arrows. The pink-dashed-pink arrow converges to the origin, and the cyan-dashed-pink arrow diverges from it. As the origin is not stable, the post-impact contact mode is always \emph{sliding} regardless of the initial tangential velocity.  
  }
  \label{fig:post_impact_vel}
  \vspace{-3mm}
\end{figure}

\subsection{Contact mode and  pre-impact tangential velocities}
\label{sec:sliding_and_tv}

\subsubsection{Non-zero pre-impact tangential velocities}
\label{sec:non-zero-preImpact-tangential-vel}
Despite \eqref{eq:zImpulseCE}  cannot predict the  tangential impulse,
the prediction $\proposedSolution$  is robust to pre-impact tangential velocities in our experiments when the friction coefficient $\coefF = 0.1123$.

Given
$\xContactVel = 0$\unitVelTS, $\nContactVel = 0.12$\unitVelTS, and 
various $\yContactVel$,  we show the mean $\measured{\jump \jvelocities}$ in \figRef{fig:tangentialVel}.
In the rightmost figure of \figRef{fig:average_error}, the average prediction errors \eqref{eq:relative_error} remain at the same level as the other experiments.

\subsubsection{Sliding contact}
\label{sec:non-zero-tangential-vel-reason}

We compute the  post-impact tangential velocities $\postImpact{\contactVel}_x, \postImpact{\contactVel}_y$ using the measured joint velocity jump $\measured{\jump \jvelocities}$ and the kinematics \eqref{eq:kinematics}. 
Given $\postImpact{\contactVel}_x, \postImpact{\contactVel}_y$ are non-zero in \figRef{fig:post_impact_vel}, the contact mode was \emph{sliding} at and after the impact.
The robot is able to slide because the manipulator is mounted  on a fixed-base, during the impact event the joint motion is compliant to  the rigid wall despite the robot is high-gain controlled.

For impacts between two free-flying objects, sliding happens if the friction coefficient is smaller than the \emph{stick coefficient} $\bar{\coefF}$ \cite[Eq.~4.12]{stronge2000book} \cite[Sec.~5]{jia2017ijrr}, i.e., the friction is not enough to sustain the contact.
Using the joint configuration in \figRef{fig:contact_posture_two} as an example, we have:
\quickEq{eq:small_coefF}{
\coefF = 0.1123 \leq \bar{\coefF} = 0.1895,
}
where $\bar{\coefF}$ is computed by substituting $\iimcrb$ into \cite[Eq.~4.12]{stronge2000book}.

Assuming the robot is a composite-rigid body, we visualize the tangential-velocity vector field (phase plane) on the right side of \figRef{fig:post_impact_vel} according to \cite{jia2017ijrr}. 
As the origin is not stable, the post-impact contact mode would be sliding regardless of the initial tangential velocities.

\section{Conclusion}
Predicting joint velocity jumps is paramount to achieve the highest feasible contact velocity while fulfilling the hardware resilience bounds. 
Our approach (i) uses our improved normal impulse calculation~\cite{wang2022ral}; (ii) avoids inverting the ill-conditioned JSIM by formulating momentum conservation equations in task space; and (iii) modifies the task-space impulse-to-velocity mapping to account for a high-gain (stiff) joint behavior in pre- and at impact, while allowing flexibility in post-impact. As a result, from benchmarking 250 impact experiments our proposed prediction $\proposedSolution$ \eqref{eq:proposed_projection} reduces the average prediction error by 81.98$\%$ w.r.t the existing approach.

Future work is dedicated to tangential impulse calculation and impact propagation in robotic structures.

\section*{Acknowledgment}
This work is supported by the EU H2020 research grant GA 871899, I.AM. project. We would like to thank Mohamed Djeha and Saeid Samadi for the valuable feedback.

\begin{figure*}[hbtp]
  \vspace{-3mm}
  \includegraphics[width=0.33\textwidth]{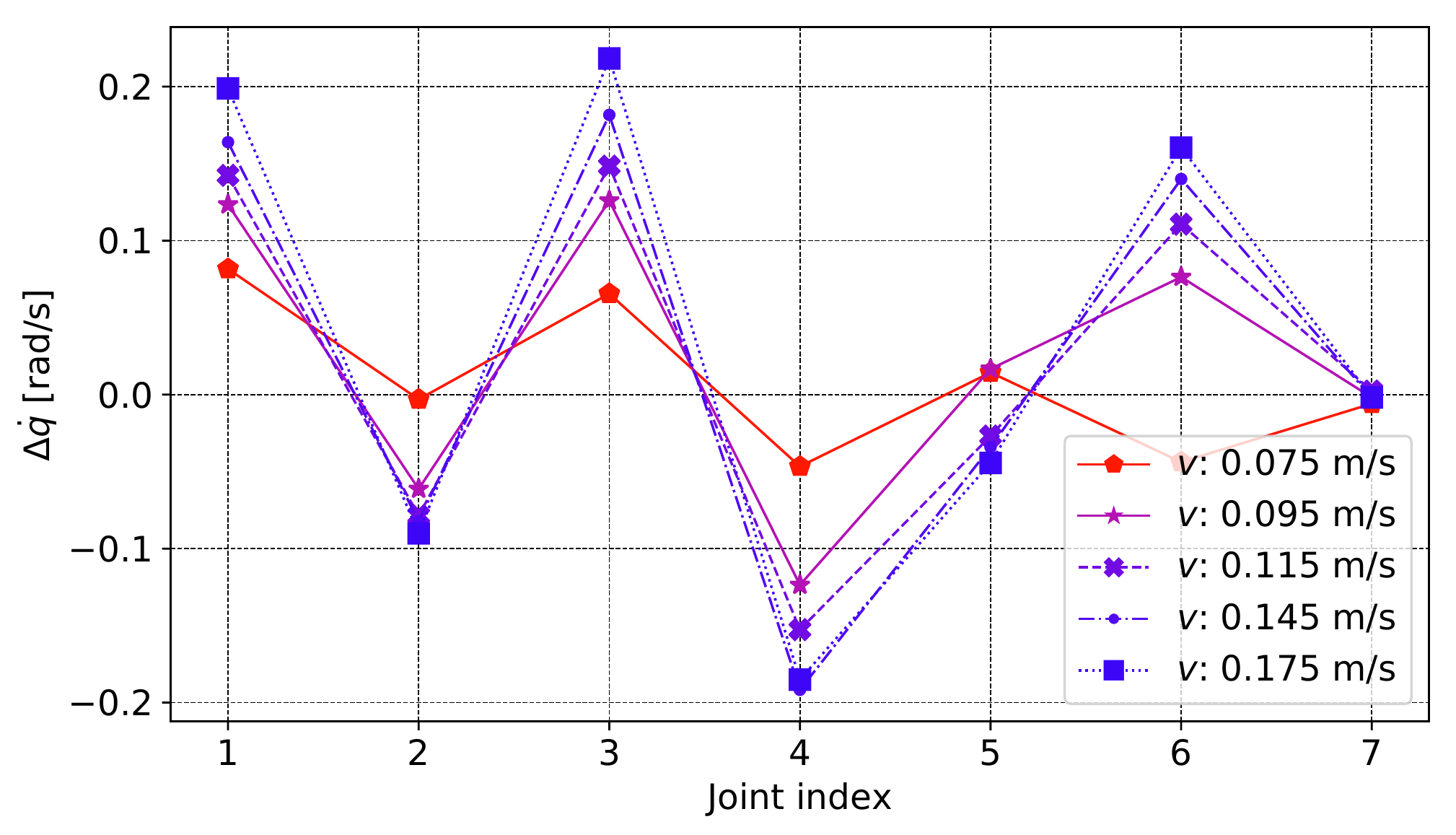}
  \includegraphics[width=0.33\textwidth]{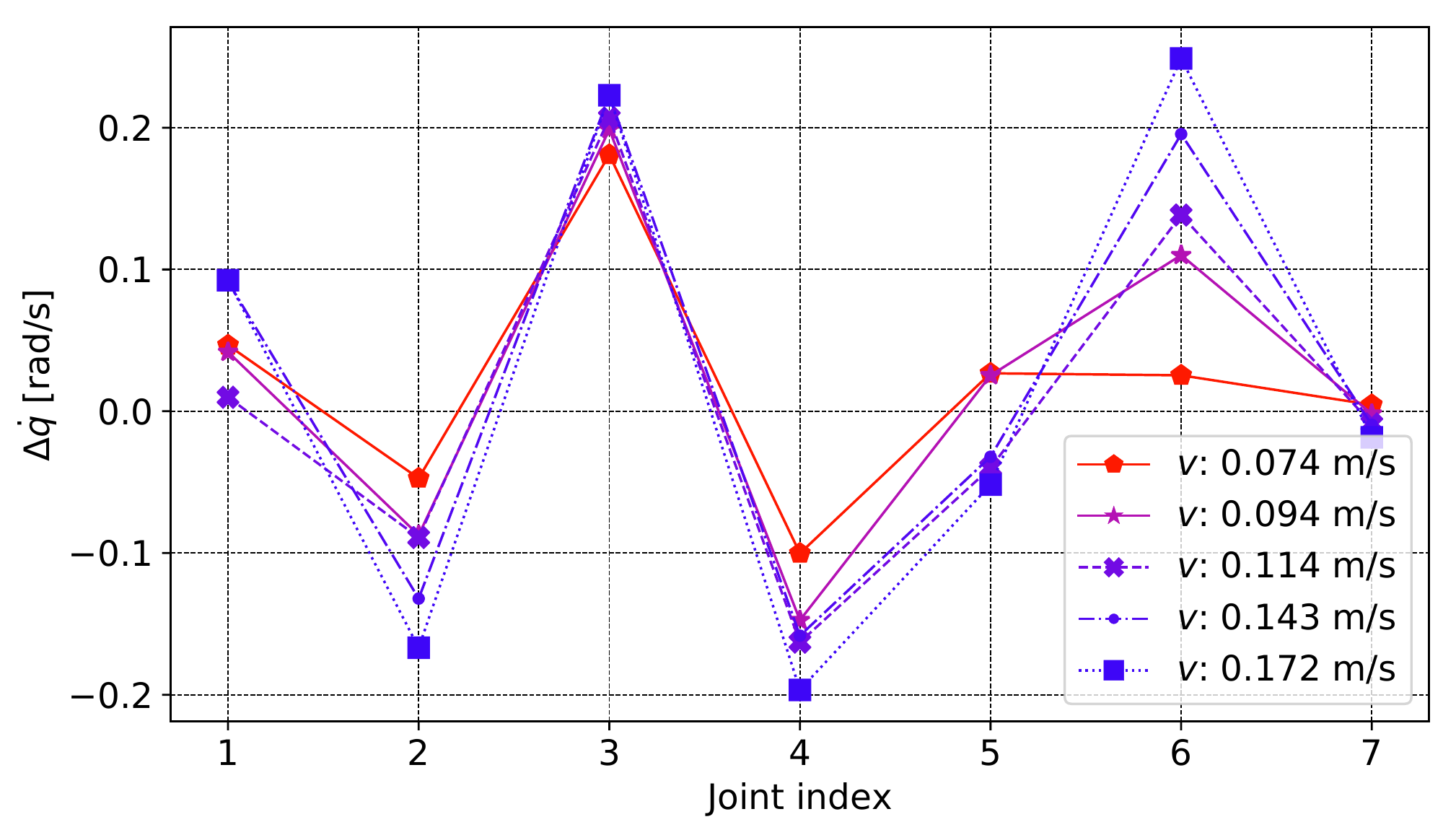}
  \includegraphics[width=0.33\textwidth]{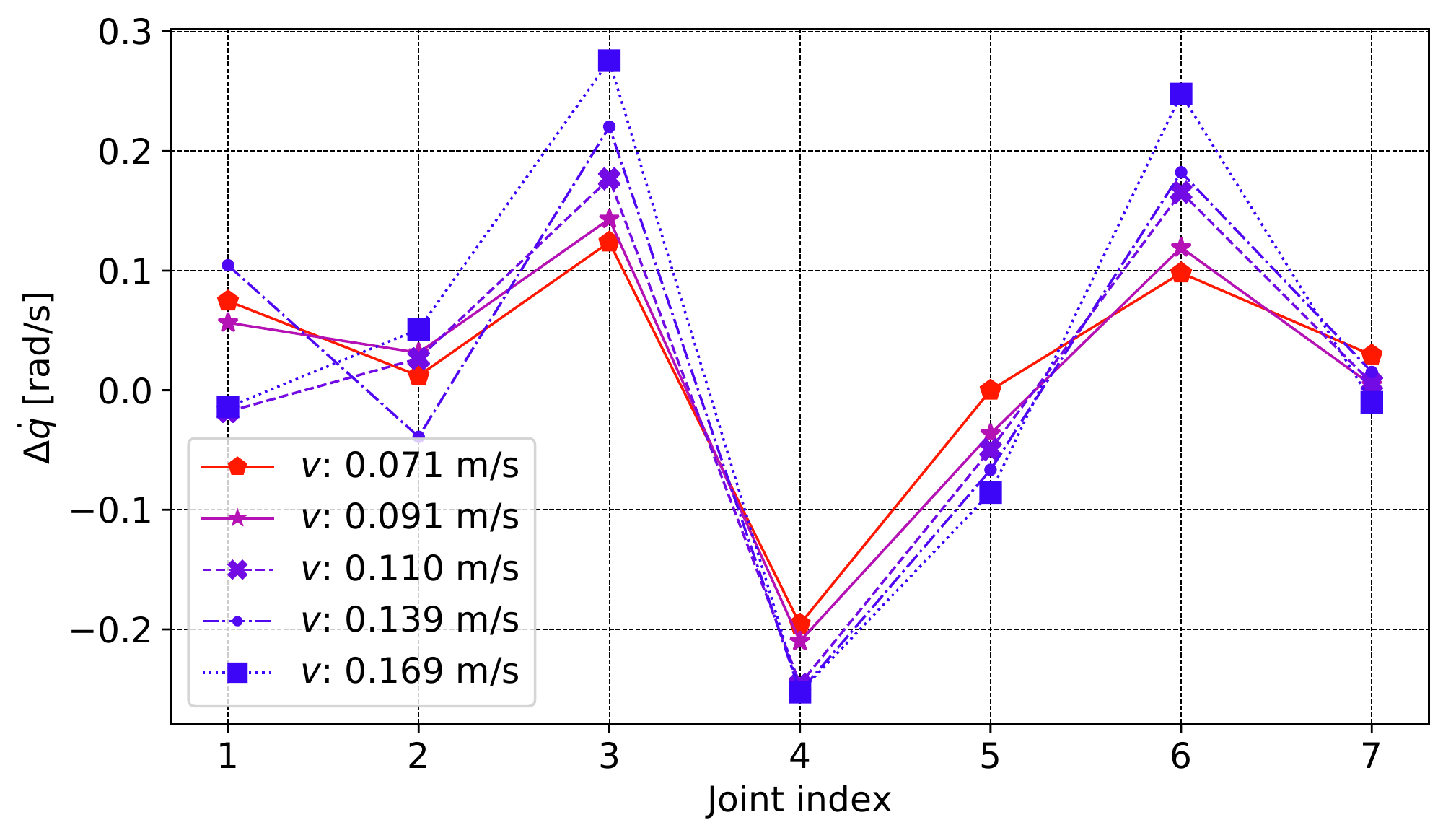}
\caption{
  From left to right, we plot the measured joint velocity jump $\jvelocitiesJumpEstimate$ of the impact configurations Fig.~\ref{fig:contact_posture_one}, \ref{fig:contact_posture_two}, \ref{fig:contact_posture_three}. Each data point denotes the mean   $\jvelocitiesJumpEstimate$ of 10 impact experiments with the same joint configurations and reference contact velocities.
}
\label{fig:qd_jump_mean}
\vspace{-3mm}
\end{figure*}
\begin{figure*}[hbtp]
  \includegraphics[width=0.33\textwidth]{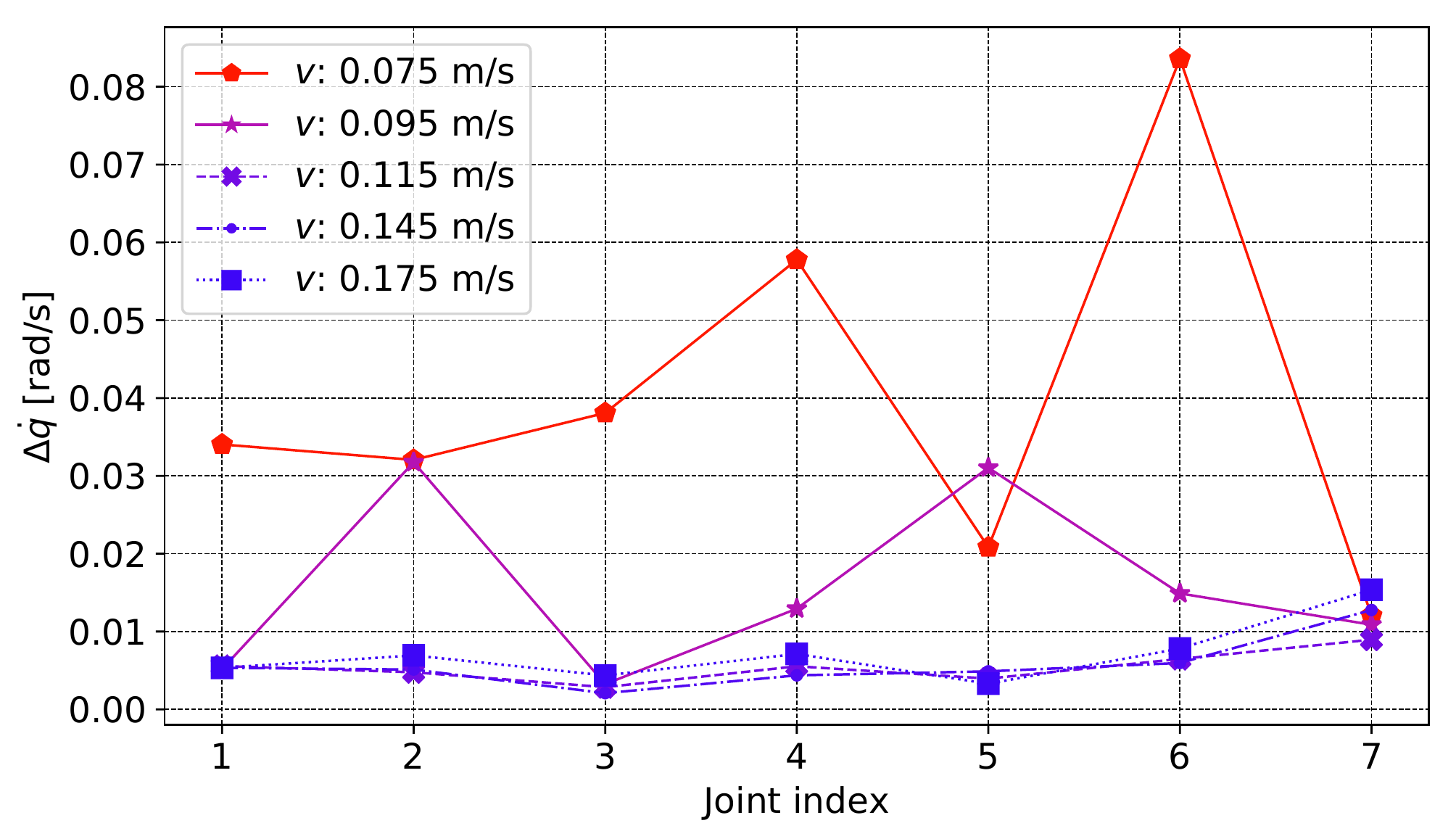}
  \includegraphics[width=0.33\textwidth]{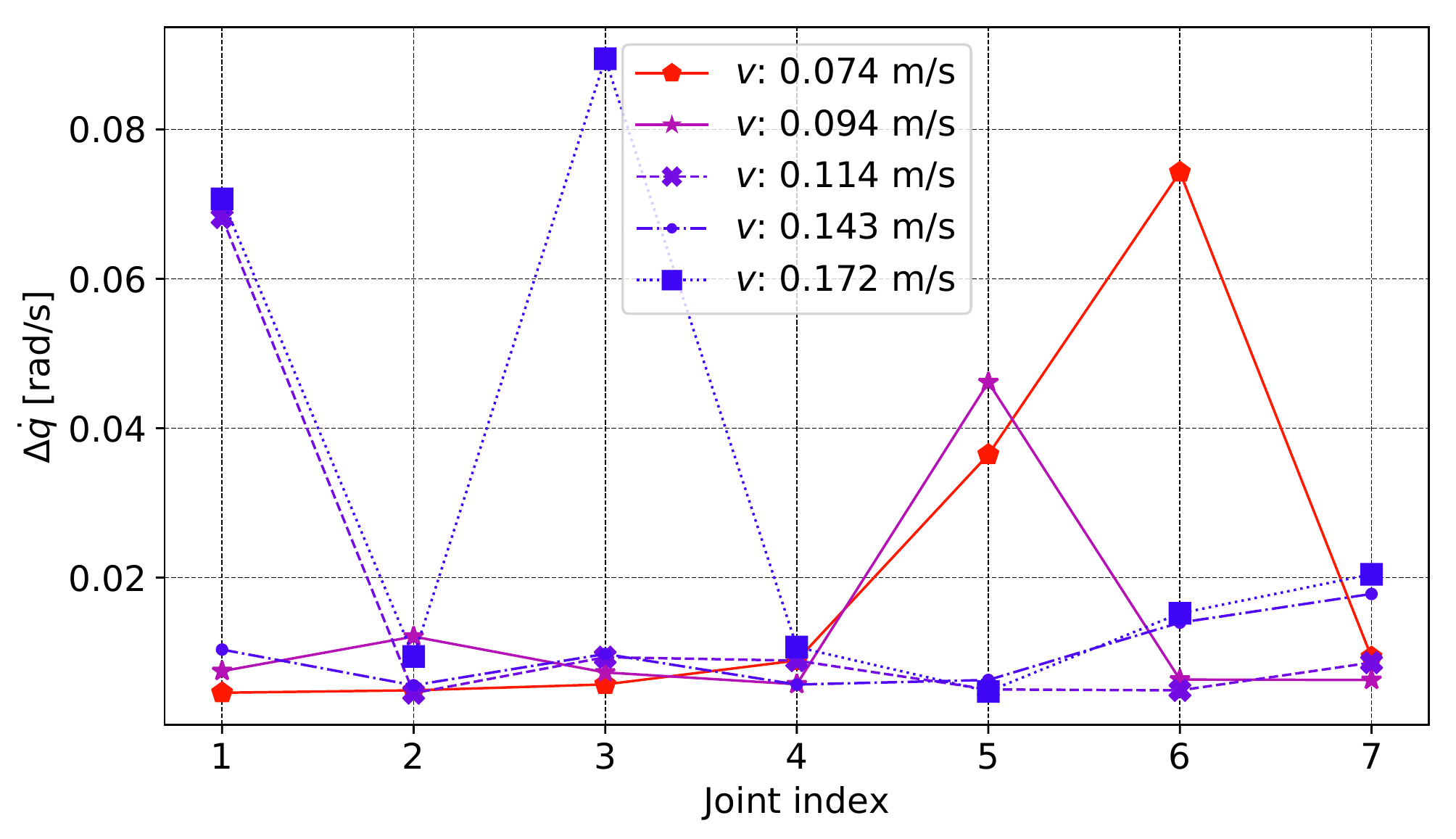}
  \includegraphics[width=0.33\textwidth]{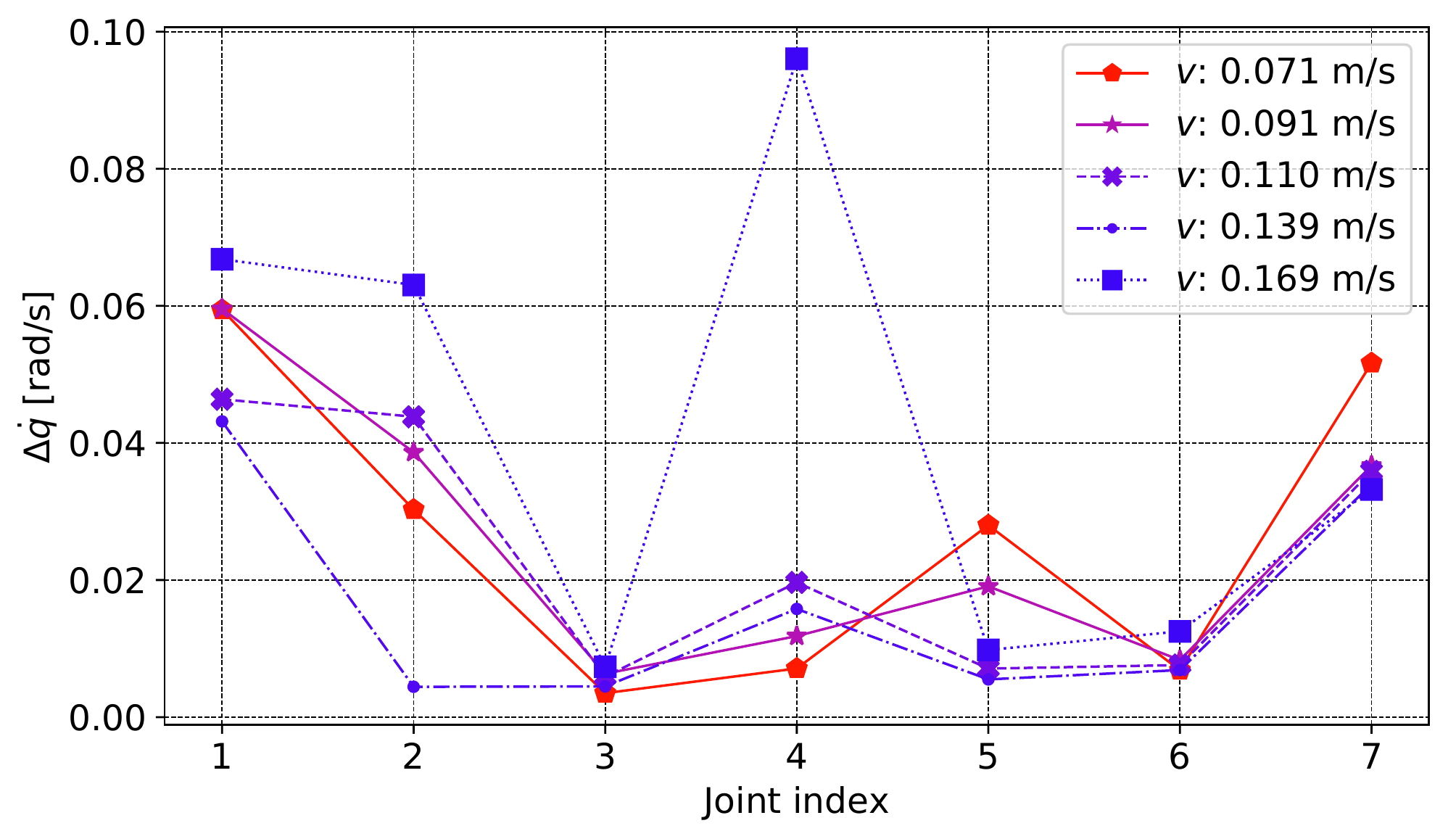}
\caption{
The standard deviations of the measured joint velocity jump $\jvelocitiesJumpEstimate$ from the impact configurations Fig.~\ref{fig:contact_posture_one}, \ref{fig:contact_posture_two}, \ref{fig:contact_posture_three}. 
}
\label{fig:qd_jump_std}
\vspace{-3mm}
\end{figure*}
\begin{figure*}[hbtp]
  \includegraphics[width=0.33\textwidth]{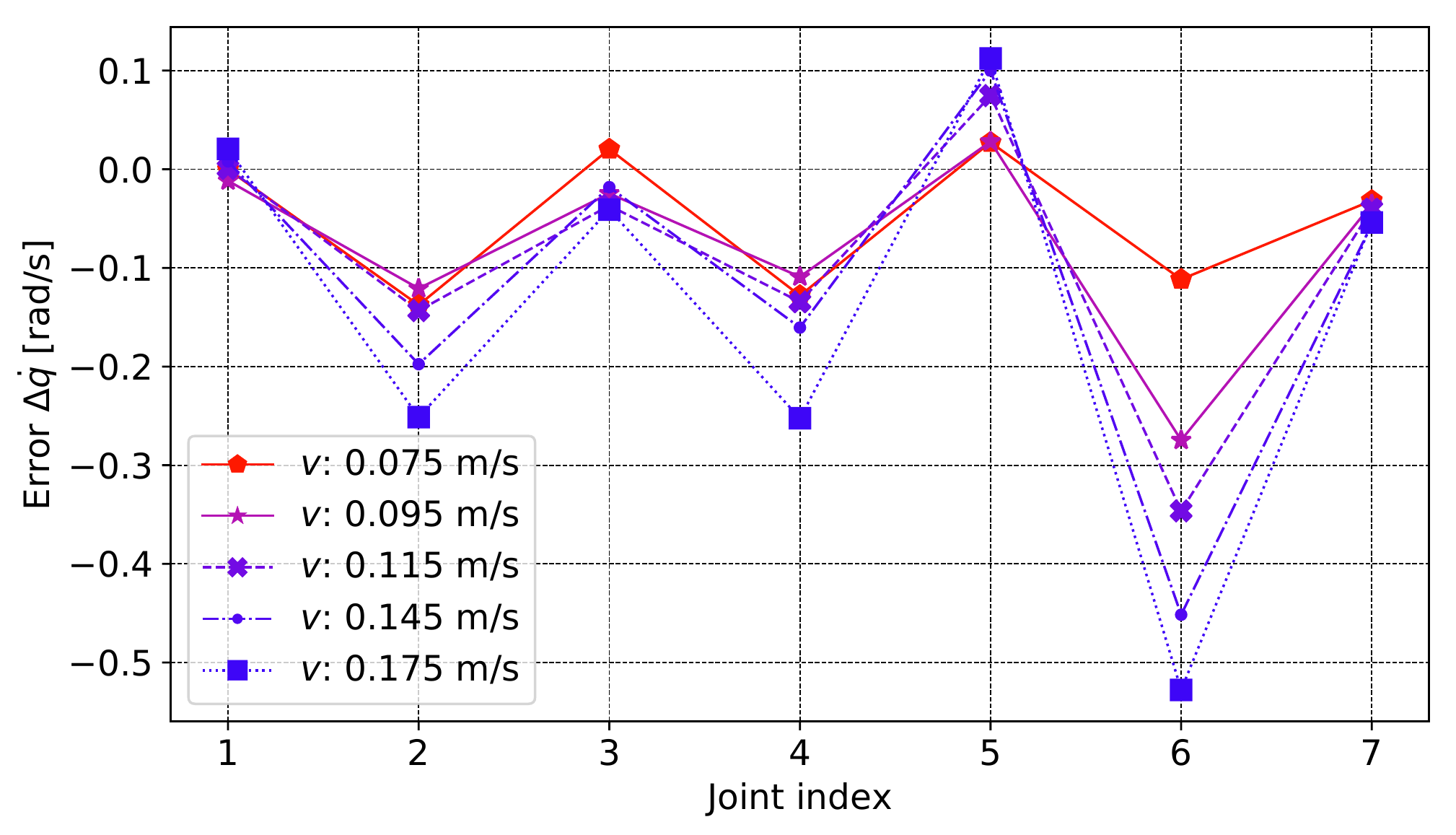}
  \includegraphics[width=0.33\textwidth]{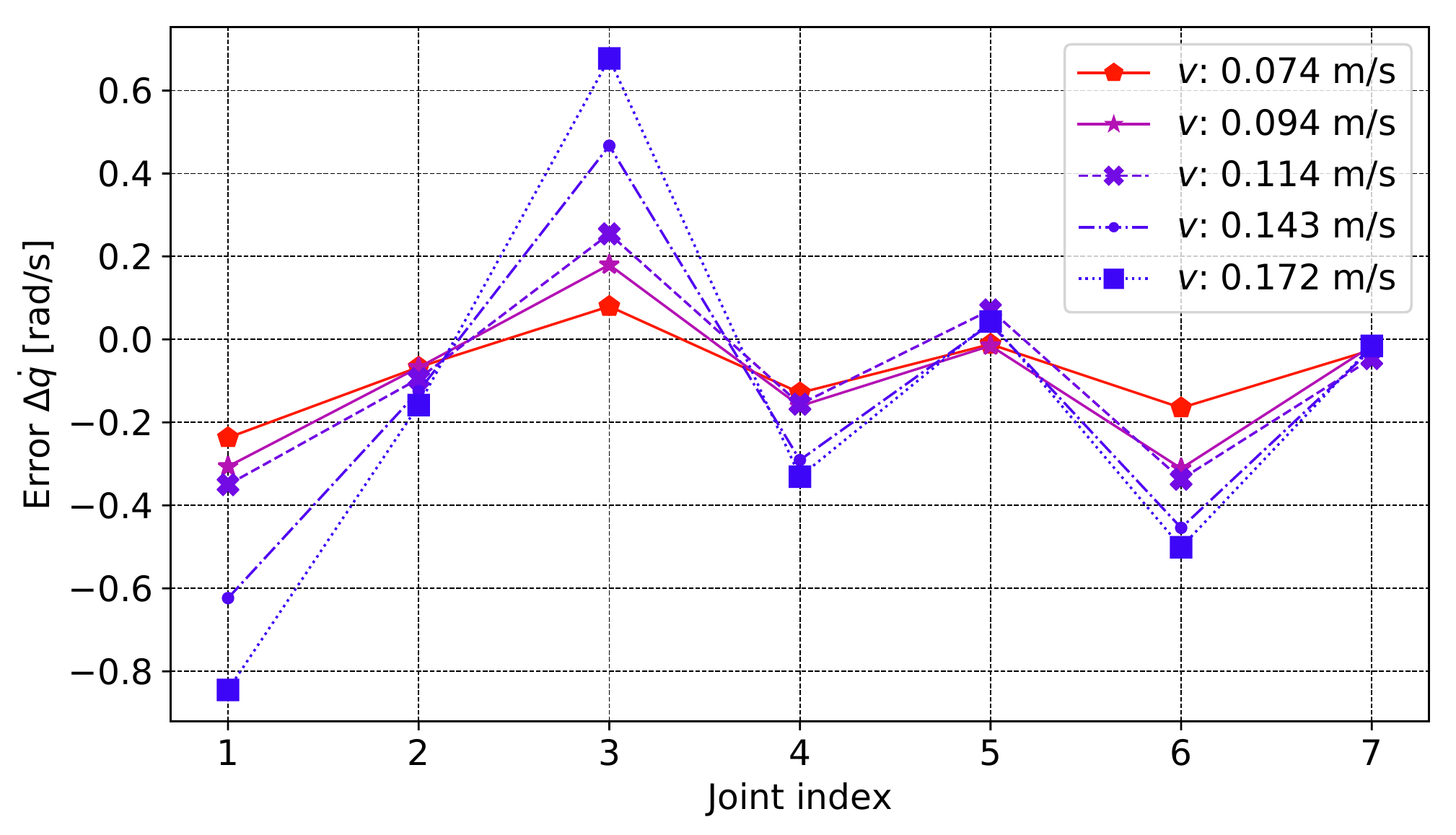}
  \includegraphics[width=0.33\textwidth]{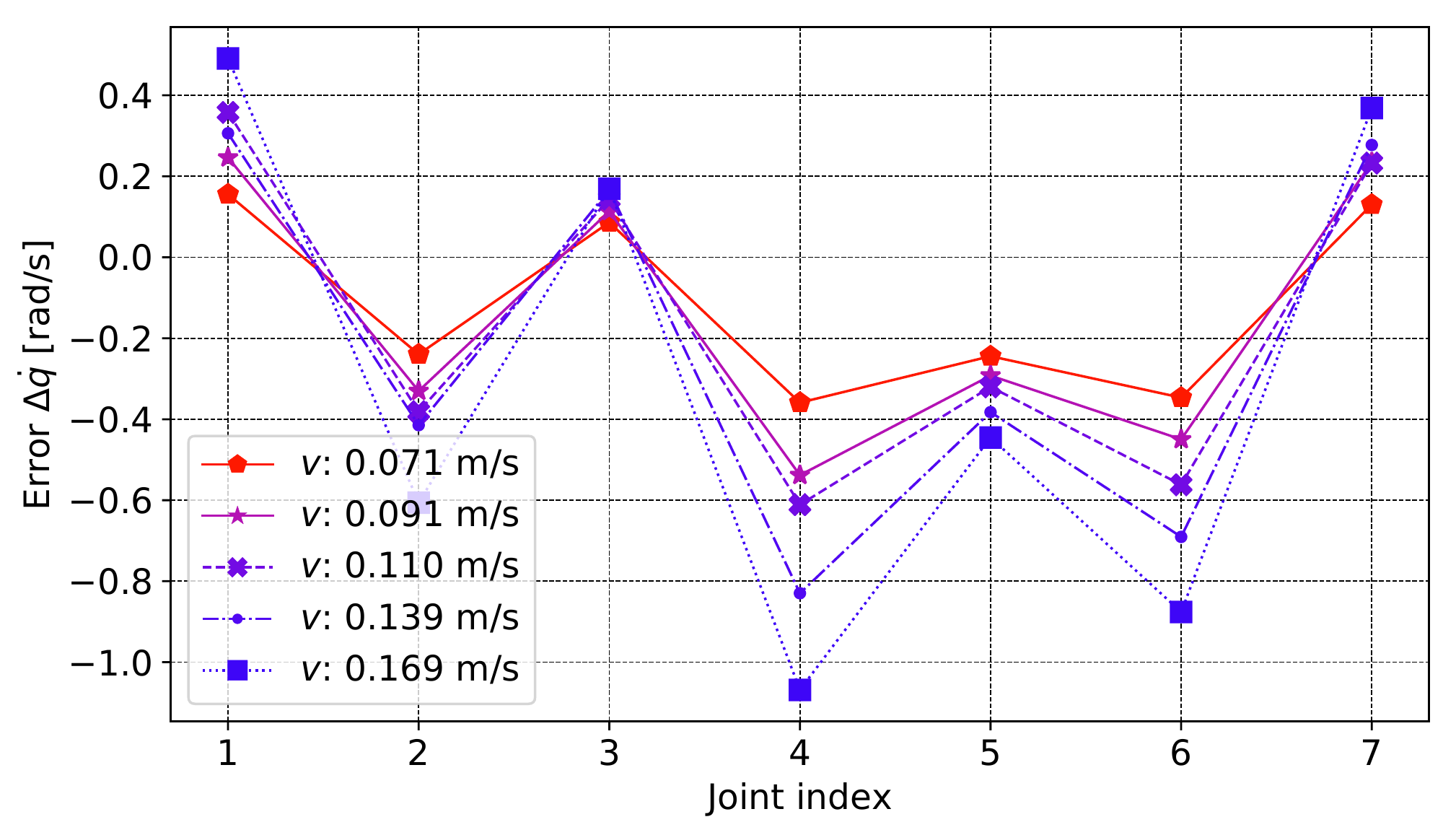}
\caption{
  Absolute prediction error of the classical solution $\classicalSolution$ \eqref{eq:textbook_dq_jump}. 
}
\label{fig:qd_jump_error_textbook}
\end{figure*}
\begin{figure*}[hbtp]
  \vspace{-3mm}
  \includegraphics[width=0.33\textwidth]{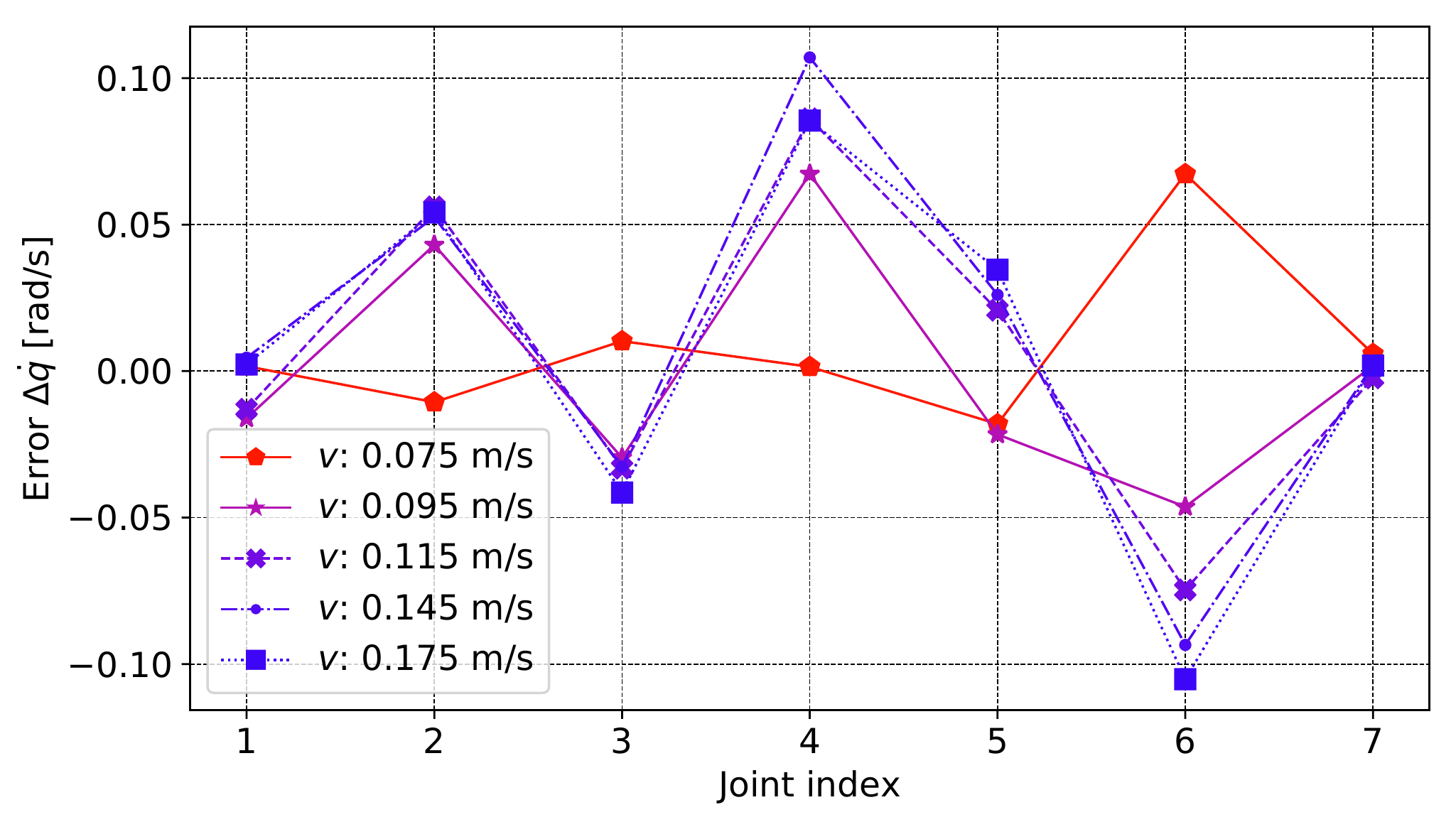}
  \includegraphics[width=0.33\textwidth]{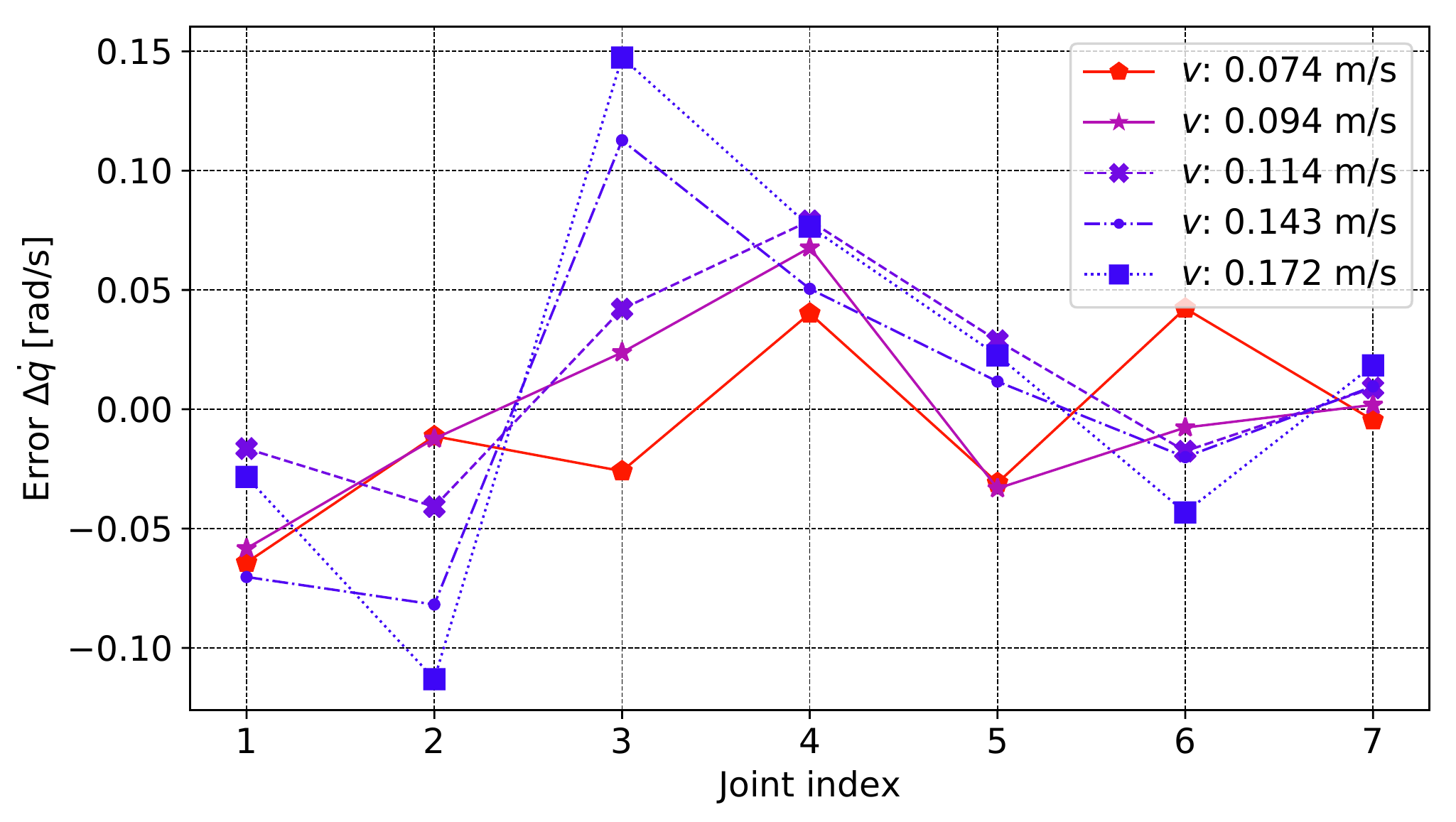}
  \includegraphics[width=0.33\textwidth]{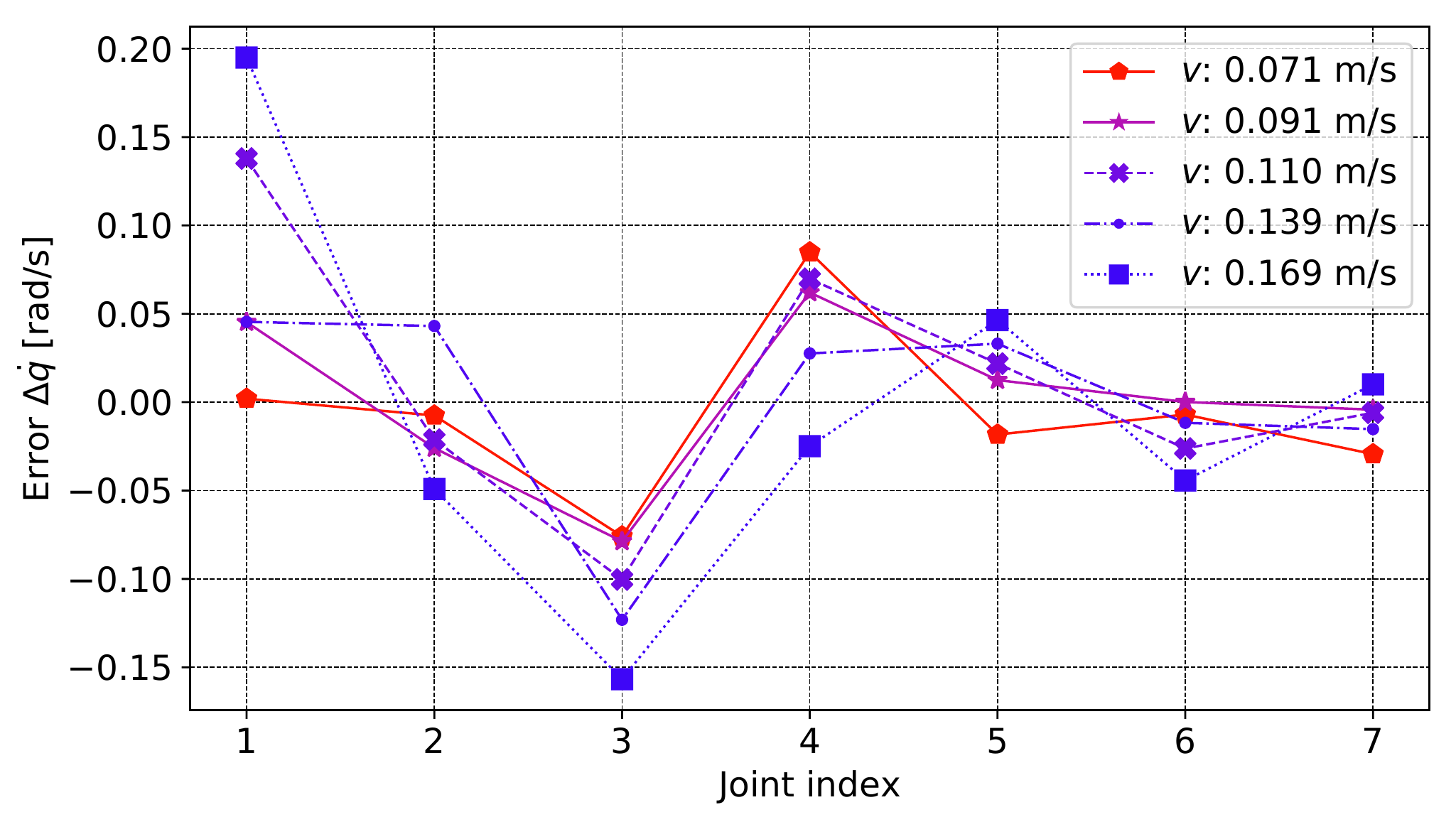}
\caption{
  Absolute prediction error of the proposed solution $\proposedSolution$ \eqref{eq:proposed_projection}. 
}
\label{fig:qd_jump_error_proposed}
\vspace{-3mm}
\end{figure*}

\begin{figure*}[hbtp]
  \includegraphics[width=0.33\textwidth]{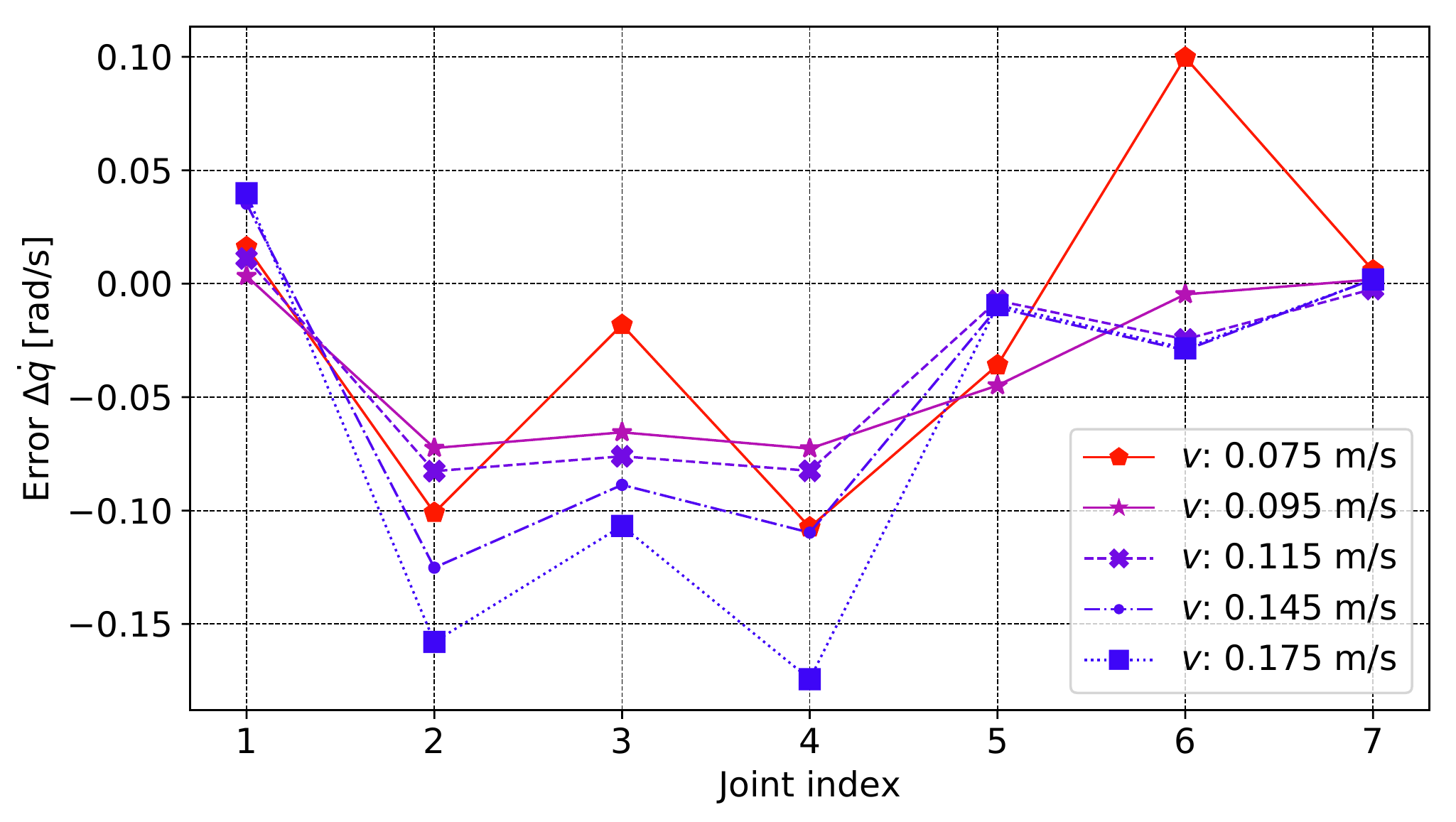}
  \includegraphics[width=0.33\textwidth]{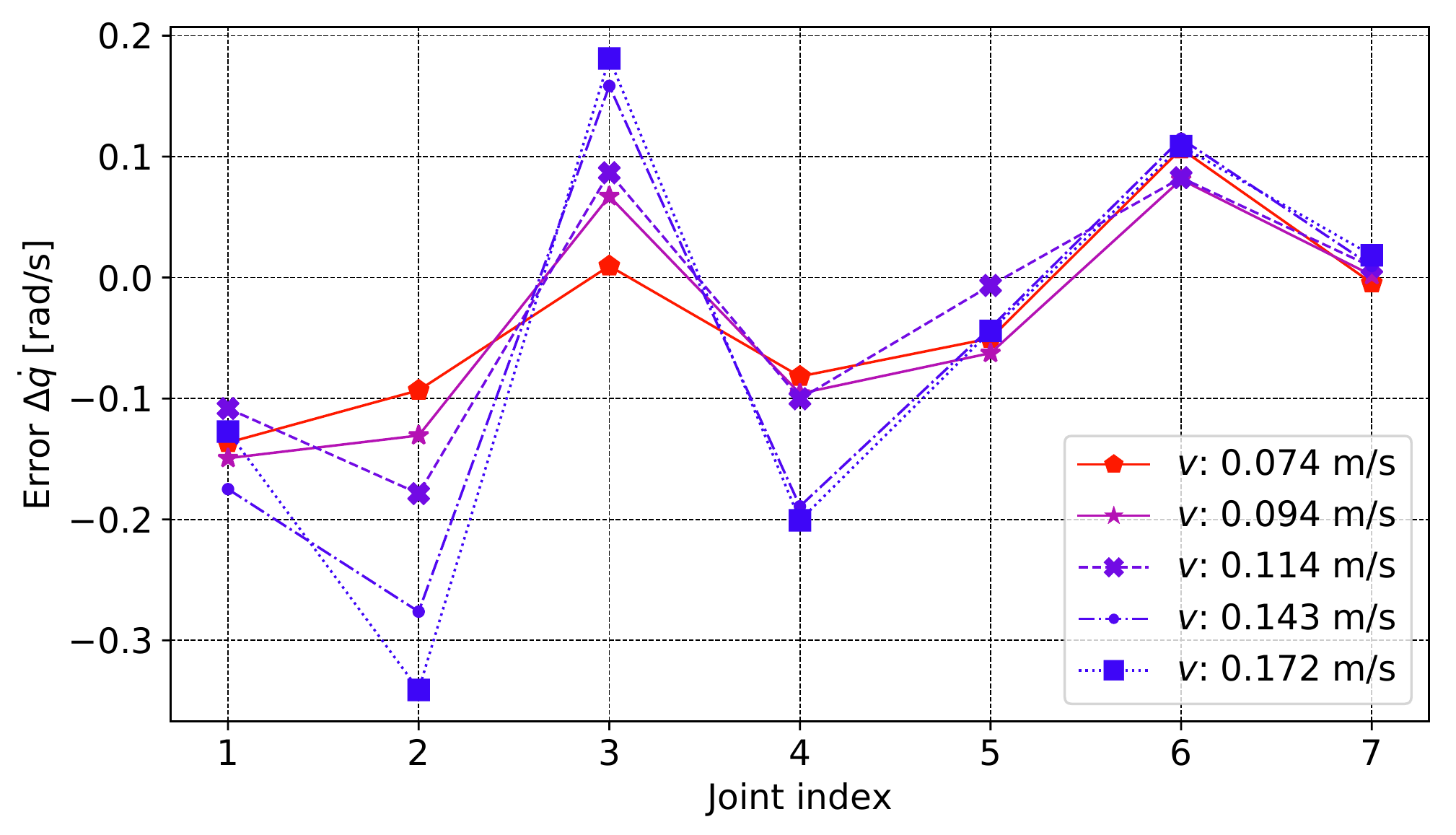}
  \includegraphics[width=0.33\textwidth]{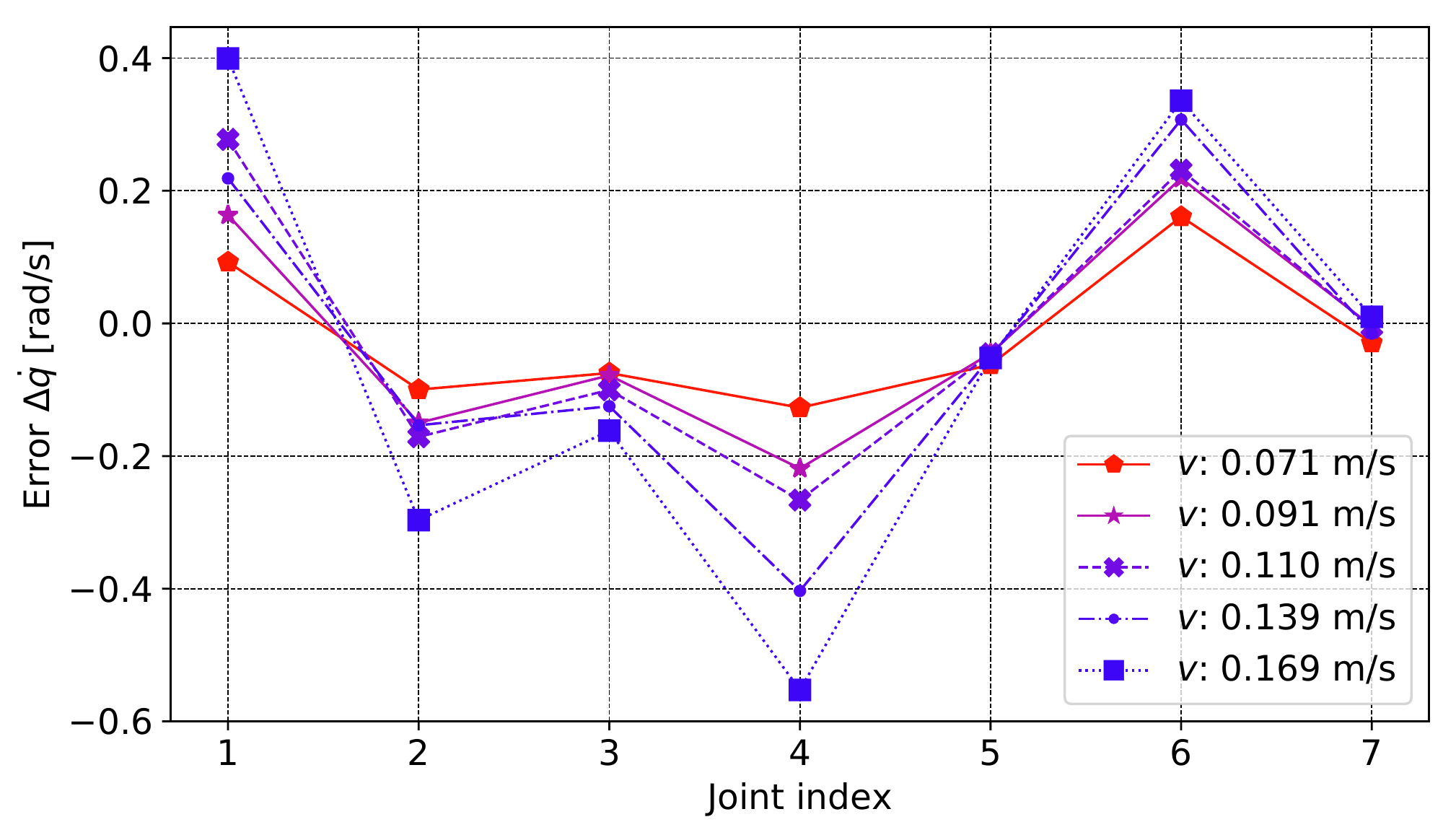}
\caption{
  Absolute prediction error of the solution $\gmSolution$ \eqref{eq:gm_projection} without the CRB assumption.
}
\label{fig:qd_jump_error_gm}
\vspace{-3mm}
\end{figure*}

\bibliographystyle{IEEEtran}
\bibliography{ref}
\end{document}